\documentclass[preprint,authoryear,12pt]{elsarticle}

\usepackage{amssymb}
\usepackage{graphicx,array,colortbl,color,rotating,amsmath,amssymb,afterpage}
\usepackage{multirow}
\usepackage{epsfig}
\usepackage{enumerate}
\usepackage{hyphenat}
\usepackage{verbatim} 
\usepackage{url}
\usepackage{multicol}

\newdefinition{example}{Example}

\usepackage{color}
\newcommand\comJ[1]{}
\newcommand\comG[1]{}
\newcommand\comL[1]{}
\newcommand\comS[1]{}

\DeclareMathOperator*{\argmax}{argmax}

\journal{}

\begin{document}

\begin{frontmatter}

\title{An Experimental Study of Adaptive Control for Evolutionary Algorithms}

\author{Giacomo di Tollo}
\address{LERIA, University of Angers (France)}
\author{Fr\'ed\'eric Lardeux}
\address{LERIA, University of Angers (France)} 
\author{Jorge Maturana}
\address{Instituto de Inform\'atica, Universidad Austral de Chile (Chile)}
\author{Fr\'ed\'eric Saubion}
\address{LERIA, University of Angers (France)}

\begin{abstract}
The balance of exploration versus exploitation (EvE) is a key issue on evolutionary computation. In this paper we will investigate how an adaptive controller  aimed to perform Operator Selection can be used to dynamically manage the EvE balance  required by the search, showing that the search strategies determined by this control paradigm lead to an improvement of solution quality found by the evolutionary algorithm.
\end{abstract}


\end{frontmatter}

\section{Introduction}

During the past decades, Evolutionary Algorithms  (EAs)\citep{Holland1975,Goldberg1989,eiben03intro} have been successfully applied to many optimization problems. From a high level point of view, EAs  manage a set of potential solutions of a problem -- a population  of individuals according to the evolutionary metaphor. The population is progressively modified by variation operators in order to converge to an optimal solution with regards to a fitness function, which evaluates the  quality of the individuals. Two well-known concepts are commonly used  to describe the behavior of an EA: \emph{exploitation} -- which reflects the ability of the algorithm to converge to an optimum -- and \emph{exploration} -- which insures that the algorithm is able to visit sufficiently sparse areas of the search space. The  balance between exploration and exploitation (referred to as EvE)   is widely recognized as a key issue of the overall search performance. This balance often relies on the adjustment of several parameters (e.g., size of the population and application rates of the different operators).

Significant progress has been achieved in parameter setting \citep{Lobo2007}. Following the taxonomy proposed by \citet{eiben99parameterControl},  \emph{tuning} techniques adjust the parameters of the algorithm before the run while \emph{control} techniques modify the behavior of the algorithm during the search process. Efficient \emph{tuning} methods are now available using    statistical tools such as racing techniques \citep{BSP02} or meta-algorithms that explore the parameters' space  (e.g., ParamILS \citep{HutHooStu07} or Revac \citep{Nannen2008}). \emph{Control} techniques   have also been proposed in order to provide adaptive or self-adaptive EAs \citep{eiben06parameterControl}. 

In this paper, we focus on Adaptive Operator Selection (AOS) methods \citep{Maturana2012} from the \emph{control} point of view: the operator selection problem consists in selecting, out of
 a set of available operators,  which one should be applied at a given iteration of the evolutionary process. The aim of AOS is to control the EvE balance in order to improve search efficiency.   Nevertheless, in most of the related works (see section \ref{previousworks}), the control of the EvE balance has been only partially investigated. 
Most of the approaches focus on exploitation and use the  quality of the population as a unique criterion to guide the search \citep{thierens07operatorAllocation,Gong2010,bandit-PPSN08}.

Furthermore, there are a few works that use several criteria to assess the utility of the operators\citep{compass-PPSN08}, but in these works the EvE balance is kept fixed. Since it has been shown that an efficient algorithm requires different parameter values during the search for achieving better results \citep{linhares_search_2010}, the EvE balance should be dynamically controlled.

The  purpose of our work is twofold. Firstly, we investigate the management of more dynamic control strategies. The framework proposed by \citet{Maturana2010} is used to implement a generic {\em controller}.\footnote{In this paper, we call {\em controller}, the complete architecture that allows us to perform adaptive operator selection.}  This controller must thus identify the suitable operators at each step of search in order to achieve the required EvE balance, which may change dynamically according to a given control strategy. Then we want to assess the impact of dynamic control on the  performance of the algorithm. Our experimental methodology is organized as follows:
\begin{enumerate}
\item {\bf Assessing the operators management}: 
 \begin{itemize}
\item by assessing whether the controller is able to identify the required operators in presence of  non-efficient operators, i.e., in presence of {\bf noisy operators};
\item 
by checking whether  the  controller is able to manage a policy in which the desired EvE balance is modified along the search. 
\end{itemize}
\item {\bf Evaluating the solving performances}:  
\begin{itemize}
\item by checking whether  the controlled EA is  able to  solve problems efficiently with regards to existing algorithms.
\end{itemize}
\end{enumerate}

\noindent In this paper we recall the main literature on the topic on section \ref{previousworks} before describing 
the controller in section \ref{sec:controller}. Then, we introduce the experimental setting in section 
\ref{experimentalSetting} before discussing  results obtained through the experimental phase: 
section \ref{opman} focuses on the management of the operators, and solving performance is investigated in section \ref{sec:performances}.

\section{Related Works}
\label{previousworks}
\comG{We are still using two tenses here. In the last email FredL proposed to use the present. Shall we stick to present? I haven't modified anything so far.} \comS{That's OK in fact}

Parameter setting \citep{Lobo2007,Eiben2012} is an important challenge for building efficient and robust EAs. As mentioned in the introduction, using an EA requires us to define its basic structural components and to 
set the values of its behavioral parameters. The components may be considered 
as structural parameters of the algorithm. Therefore, parameter setting in EA addresses two general classes of parameters: \emph{structural} and  \emph{behavioral} (alternatively, the terms \emph{numerical} and \emph{symbolic} parameters are used \citep{smit09comparing}). Concerning structural parameters, automated tuning techniques \citep{Hoos2012} can be used as  tools for selecting the initial configuration of the algorithm. The configuration and the discovery of new heuristics from building blocks  is also addressed by the concept of hyperheuristics \citep{Burke2009a}.  We may also mention self-adpative operators  that mainly consists in encoding directly the parameters of the oprator in the individuals. This approach also allows the algorithm to dynamically manage the EvE balanceand has been successfully applied for solving combinatorial and continous optimization problems \cite{Zhang2009,Tang2014,Qin2009,Tang2013}. Note that an adaptive management of the operators, which dynamically adds and discards operators during the search, has been proposed by \citet{Maturana2010}.

In this paper, we focus on behavioral parameters and we limit our investigation to the Adaptive Operator Selection. 
AOS can be seen as the choice of the best policy  for selecting the operators during the search and different methods have been proposed to this goal.

Let us consider n operators: the probability of selecting operator $op_i$ at time $t$ is $s_i(t)$. In a static setting, the probability of selecting $op_i$ (for each i) is fixed over time (i.e., $s_i(t) = s_t(t')$, for any  $t$ and $t'~ \in~ [1, t_{max}]$ ), and can be determined by an automated tuning process. Contrary to a static tuning of the operator application rates, adaptive operator selection consists in selecting the next operator to apply at time $t+1$ by adapting the selection probability during the search  according to the performance of the operators. Let us consider an estimated utility $u_i(t)$ of operator $op_i$ at time t. This utility of the operators has to be re-evaluated at each time, classically using a formula $u_{i}(t+1) = (1-\alpha) u_{i}(t) +\alpha r_{i}$ where $r_i$ is the reward associated to the application of operator $op_i$ (immediate performance) and $\alpha$ is a coefficient that control the balance between past and immediate performance, as done in classic reinforcement learning techniques \citep{Sutton1998}. Note that $\alpha$ can be set to $\frac{1}{t+1}$ in order to compute the mean value. A classic selection mechanism  is the probability matching selection rule (PM) and can be formulated as: 

\begin{equation}
\label{eqn:PM}
s_{i}(t+1) = p_{min} + (1-n\times p_{min}) \frac{u_{i}(t+1)}{\Sigma_{k=1}^{n} u_{k}(t+1)},
\end{equation}

\noindent where a non negative $p_{min}$ insures a non zero selection probability for all operators \citep{GoldbergMatching90,Lobo1997}. 

Thierens \citep{Thierens2005,thierens07operatorAllocation} has explored a {\em winner-take-all} strategy for AOS, based on the quality (or fitness) of the population:

\begin{equation}
\label{eqn:AOSFitnessFunction}
\left\{
\begin{array}{l}
s_{i^*}(t+1) = s_{i^*}(t) + \beta (p_{max} - s_{i^*}(t)) \\
s_{i}(t+1) = s_{i}(t) + \beta (p_{min} - s_{i}(t)) \\
\end{array},
\right.
\end{equation}

\noindent where $i^*=\argmax  \{u_{i}(t), i=1..n \}$,  $p_{max} = 1-(n-1)p_{min}$ and $\beta$ is a parameter to adjust balance of this winner-take-all strategy.

Alternatively, AOS can also be considered as amulti-armed bandit problem, which is a classic reinforcement learning problem \citep{Sutton1998}.  The initial multi-armed bandit problem was introduced in the context of the experiment design by \citep{Robbins1952}. It was formulated as the maximization of the total gain of a gambler who could make $n$ tosses with two coins $A$ and $B$ with a gain of $1$ for each head but nothing for tails. The biases of the coins are unknown. This problem is known as the {\em Two-armed Bandit} and has been extended to multi-armed bandit by Rodman \citep{Rodman1978}.  Later, Auer \citep{Auer2002b} has proposed to use this problem to manage the compromise between exploration and exploitation in optimization algorithms.
The $MAB$ (Multi-Armed Bandit) algorithms that uses an UCB (Upper Confidence Bound) in order to approximate the 
expected benefit of an operator $ op_i$ at time $t$ have been firstly extended to AOS by \citet{bandit-GECCO08}:  the operator that maximizes $Mab_{i}(t)$ in the following formula  is selected:

\begin{equation}
\label{eqn:DaCostaAOS}
Mab_{i}(t) = u_{i}(t) + C \sqrt{ \frac{\log \sum_{j\in1..n} n_{j}(t)  }{n_{i}(t)} },
\end{equation}

\noindent where $r_{i}(t)$ is the reward of operator $op_i$ at time $t$, $n_{i}(t)$ is the number of times operator $op_i$ has been applied so far, and $C$  is the scaling factor  used to properly balance rewards and application frequency.   In the initial multi-armed bandit problem, the expected gain of each possible action is supposed to be fixed over time. Therefore, in \citet{bandit-GECCO08}, the authors  propose to use a Page-Hinkley  test in order to detect a change of the behavior of the operators  and thus to reset $r_{i}(t)$ and $n_{i}(t)$.  In \citep{Fialho2010a}, an improved technique has been proposed for comparing the respective performance of the operators. 

Note that the equation \ref{eqn:DaCostaAOS} uses $n_{i}(t)$ as a way to avoid forgetting less favorable operators, supposing that all operators were included from the start of the search. Indeed, if one of them were introduced to the eligible set in the middle of the search,  it would be necessary  to apply the operator several times to catch up with respect to the other ones. This would imply a waste of time and an eventual degradation of the search if the new operator would not be suited to the current search requirements. In order to deal with this situation, a variation of the AOS was proposed in \citep{Maturana2010} that considers \emph{idle time} instead of the number of times an operator has been applied.

Focusing on the  performance measures,  \citet{Whitacre2006} consider extreme values over a few applications of the operators, based on the idea that highly beneficial but rare events might be more beneficial than regular but smaller improvements.  

Most works rely on quality as the only criterion used for control. Nevertheless, EA literature has constantly been concerned with maintaining the diversity of the population in order to avoid premature convergence \citep{div1}. 
Therefore,  \citet{compass-PPSN08} have proposed
another AOS method, which  manages simultaneously the mean quality and the diversity of the population: these two criteria are clearly related to the exploitation and the exploration of the search process. The impact of an operator is thus recorded in two sliding time windows and used to select the next operator according to a given search trajectory, which is defined in this two-dimensional performance space. 

 \citet{cec2009} have evaluated several  combinations of various control components   using  ideas from \citet{bandit-PPSN08}, \citet{bandit-GECCO08}, and \citet{compass-PPSN08}. These works have investigated different   methods for rewarding operators according to their performances, and different operator selection techniques for choosing the most suitable operator according to its past rewards. 

In all these works the balance between these criteria, which can be seen as an abstraction of the exploration-exploitation balance, is set according to a fixed and predefined search policy. In this paper, instead, we want to explore alternate possibilities offered by this powerful AOS framework in order to provide a more dynamic management of the  algorithm's behavior with regards to this balance. 

\section{A Generic Controller for Selecting Variation Operators}
\label{sec:controller}

This section describes a generic {\em controller} for Adaptive Operator Selection (AOS) in evolutionary algorithms. In order to assess the generality of our controller,  we consider a generic EA that may include several operators.  This controller is connected to the algorithm by a simple  I/O interface: 
\begin{itemize}
\item
the EA sends the controller the identifier of last applied operator identifier and its associated performance values; 
\item
the controller tells the EA which operator should be applied next. 
\end{itemize}

\noindent
 AOS  relies on  performance criteria which are computed and received from  the EA. These criteria are meaningful measures of the utility of the applied operator over the search. In order to keep an independence from the EA, the criteria are calculated by the latter, and sent to the AOS. The specific criteria set  considered in this work is the one used in \citet{compass-PPSN08}, where two performance criteria are used to reflect the EvE balance: the mean quality (fitness) of the population and the diversity (entropy) of the population. The choice of the mean quality is rather straightforward; the choice of the entropy need  some justification. For instance, we could have used the fitness diversity or the edit distance \citep{Rivest1992} instead, but they appeared too much correlated with the fitness \citep{Burke2004}. Preliminary experiments have shown  that  entropy shows a negligible correlation with fitness when the controller aims to favor diversity. Hence, entropy provides us with a clear information on the phenotypic distribution of the population. Please notice that each time we mention the  values of the criteria, we are  interested in their variation   rather than in their absolute current values.

\begin{figure}[!ht]
\begin{center}
\includegraphics[scale=0.35]{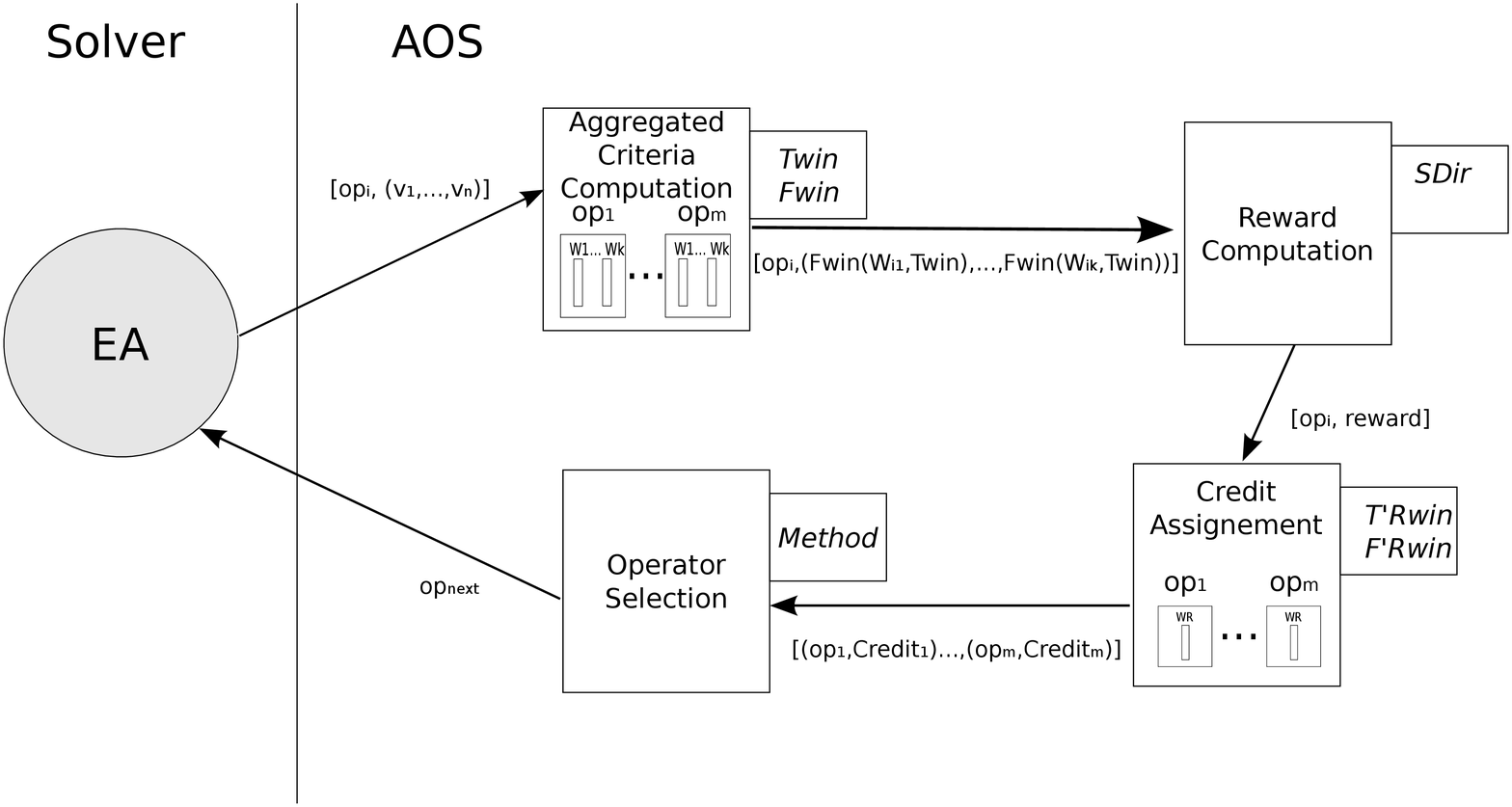}

\vspace*{0.5cm}

\begin{tabular}{|c|}
\hline
Notations \\
\begin{tabular}{|l|m{10cm}|}
\hline
$op_i$, $op_{next}$ & Operator identifiers  ($i,next~\in [1 \ldots m]$)\\
\hline
$v_j$ & Observed value of the performance criterion $j$\\ 
\hline
$Twin$ & Size of a sliding windows containing recorded data\\
\hline
$Fwin$ & Function to aggregate performance criterion values in sliding windows\\
\hline
$SDir$ & Search direction, related to performance criteria\\
\hline
$reward_i$ & Reward of  operator $op_i$ (numerical value) \\
\hline
${Credit}_i$ & Credit assigned to  operator $op_i$ w.r.t. its rewards\\
\hline
$Method$ & Operator Selection Method\\
\hline
\end{tabular}\\
\hline
\end{tabular}
 \end{center}

\caption{ AOS General Scheme}
\label{AOS:scheme}
\vspace{-0.4cm}
\end{figure}

The controller mechanism is divided into four basic stages:  \emph{Aggregation Criteria Computation}, \emph{Reward Computation}, \emph{Credit Assignment}, and \emph{Operator Selection}. These stages define a chain of modules, which are presented in figure \ref{AOS:scheme}.  Each module has its own inputs, outputs and parameters. The parameters of each module are highlighted into boxes at the right of each module.

\paragraph{\textbf{Aggregated Criteria Computation}}

This module records the  impact of the successive applications of an operator during the search. This impact corresponds to the variation of the value of the above mentioned  criteria. In  order to deal with the long-term behavior of the operators, the values are recorded in a sliding window of size $Twin$. A sliding window $W_{ij}$ is associated to each pair  $(op_i, j)$ of  operator $op_i$ and criterion $j$. The impact is then computed as the result of a function $Fwin$ applied on the window for each criterion. $Fwin$ can be instantiated to $max$ if one aims at detecting outliers, or $mean$ if one wants to smooth the behavior of the operator. 
The  input of this module are the identifier of last applied  operator ($op_i$) and the observed variation of 
the $k$ criteria values ($v_1 \ldots v_k$); the output -- sent to the \emph{Reward Computation} module -- is thus a vector  $[op_i, Fwin(W_{i1},Twin),\ldots, Fwin(W_{ik},Twin)]$. Note that at the end of ACC, only one (aggregated) scalar value is issued for each couple (operator, criteria).

\paragraph{\textbf{Reward Computation}}

\begin{figure}[!ht]
\begin{center}
  \includegraphics[width=3cm]{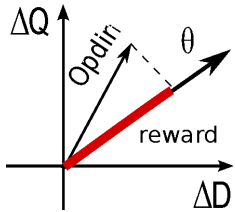}
 \end{center}
\caption{Compass Reward Computation.\label{figcompass}}
\end{figure}

Once the behavior of each operator is computed, we are interested in assessing comparatively the available operators. This comparative measure is denoted as \emph{reward}.
In this work we will use  the \emph{Compass} method \citep{compass-PPSN08}, that defines a search angle $\theta \in [0,\frac{\pi}{2}]$ in the 2-dimensional space defined in the $\Delta Diversity/\Delta Quality$ space, as shown on figure \ref{figcompass}. Each operator is thus represented in this two dimensional space according to its previous aggregated impact, and associated to a vector $opdir_i$. \comG{I would magnify the following picture (just a bit)}

\noindent A search policy is thus fully defined by the value of $\theta$: $\theta = 0$ corresponds to a policy in which the diversity is fostered and the quality is neglected; $\theta = \frac{\pi}{2}$ corresponds to  a policy in which the quality is fostered and the diversity is neglected. The reward\footnote{The term \emph{reward} is usually used in AOS methods and refers to the benefit provided by  the application of a particular action.} is computed as the scalar product  between the  vector defined by $\theta$ and $opdir_i$. 

In Compass, the angle $\theta$ stands for the variable $SDir$ in the reward computation module. However, it must be noted that other measures may be used to establish the search policy.
\footnote{For instance, \citep{Nada2012lion} proposes a method to vary the preference between two criteria in local search: quality and distance from the search trajectory. A parameter $\alpha$ controls which of these two criteria must be preferred in a Pareto-based comparison among them. In this case, $SDir$ could be mapped to the $\alpha$ parameter.}

The vector $[op_i, Fwin(W_{i1},Twin),\ldots, Fwin(W_{ik},Twin)]$ is the input of this module. Only one (aggregated) scalar value is determined for each couple (operator, criteria). The output of this module is the reward of the operator $op_i$, corresponding to its impact according to the criteria expressed as a single value.

\begin{figure}[!ht]
\begin{center}
\includegraphics[angle=-90,width=0.45\textwidth]{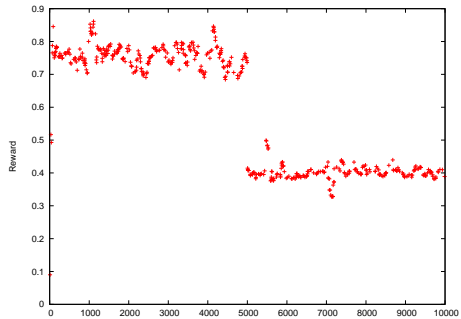}
\end{center}
\caption{Operator 0011: rewards\label{0011}}
\end{figure}

The rewards obtained by an operator will be closely related to the state of the search. Figure \ref{0011} shows the rewards obtained by an exploration operator into a context of a search strategy that encourages exploration during the first 5\,000 iterations and exploration during the remaining 5\,000. Notice how this operator is better rewarded when its behavior is coherent with the policy defined by the strategy during the first half of the search (see section \ref{Op} for more details).

\comG{Also here I would magnify the figure a bit}

\begin{figure}[h]
\begin{center}
\begin{minipage}[t]{0.45\textwidth}
\begin{center}
\includegraphics[angle=-90,width=\textwidth]{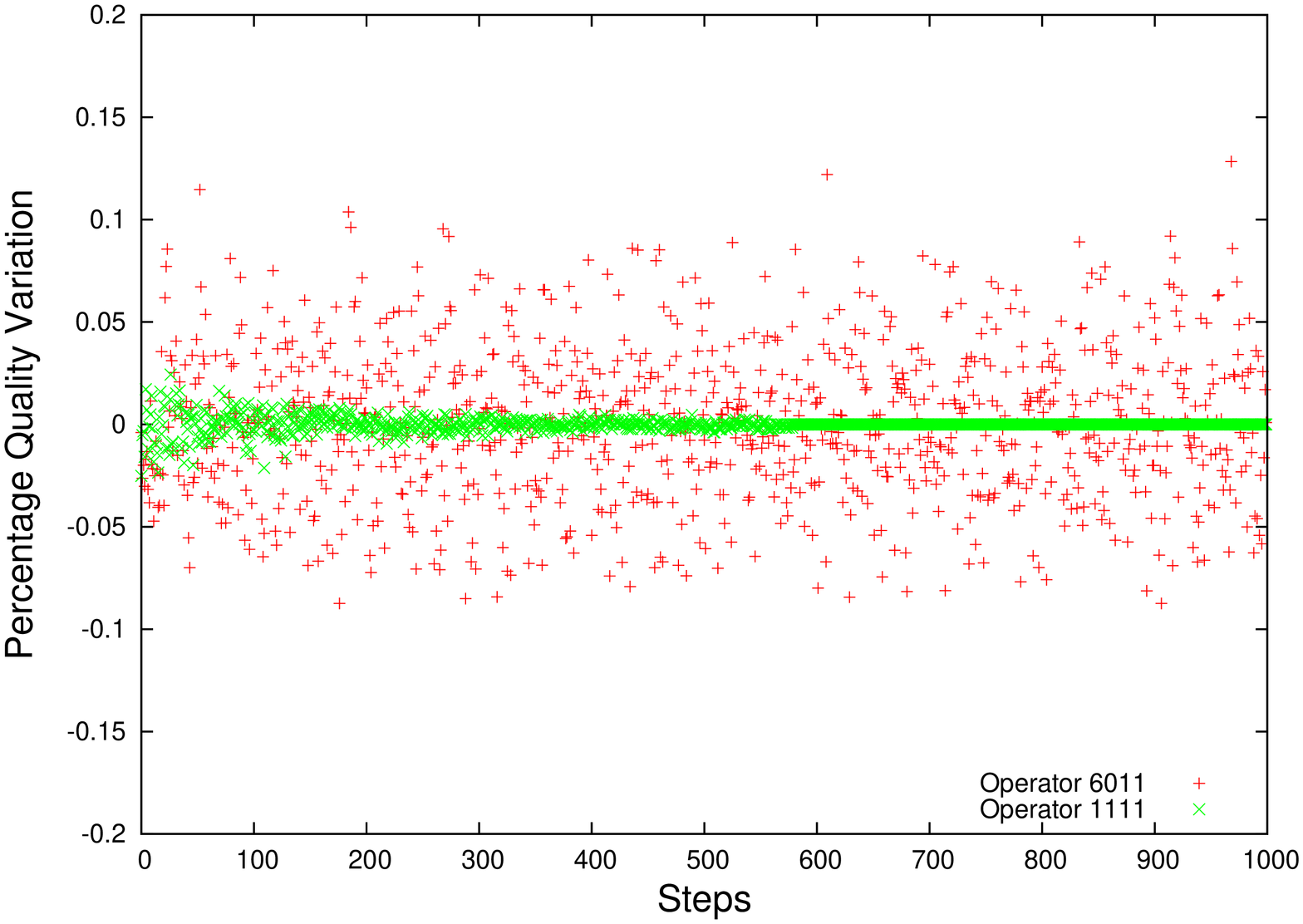}
(a) Impact on $\Delta Q$, Operators $1111$ and $6011$.
\end{center}
\end{minipage}
\begin{minipage}[t]{0.45\textwidth}
\begin{center}
\includegraphics[angle=-90,width=\textwidth]{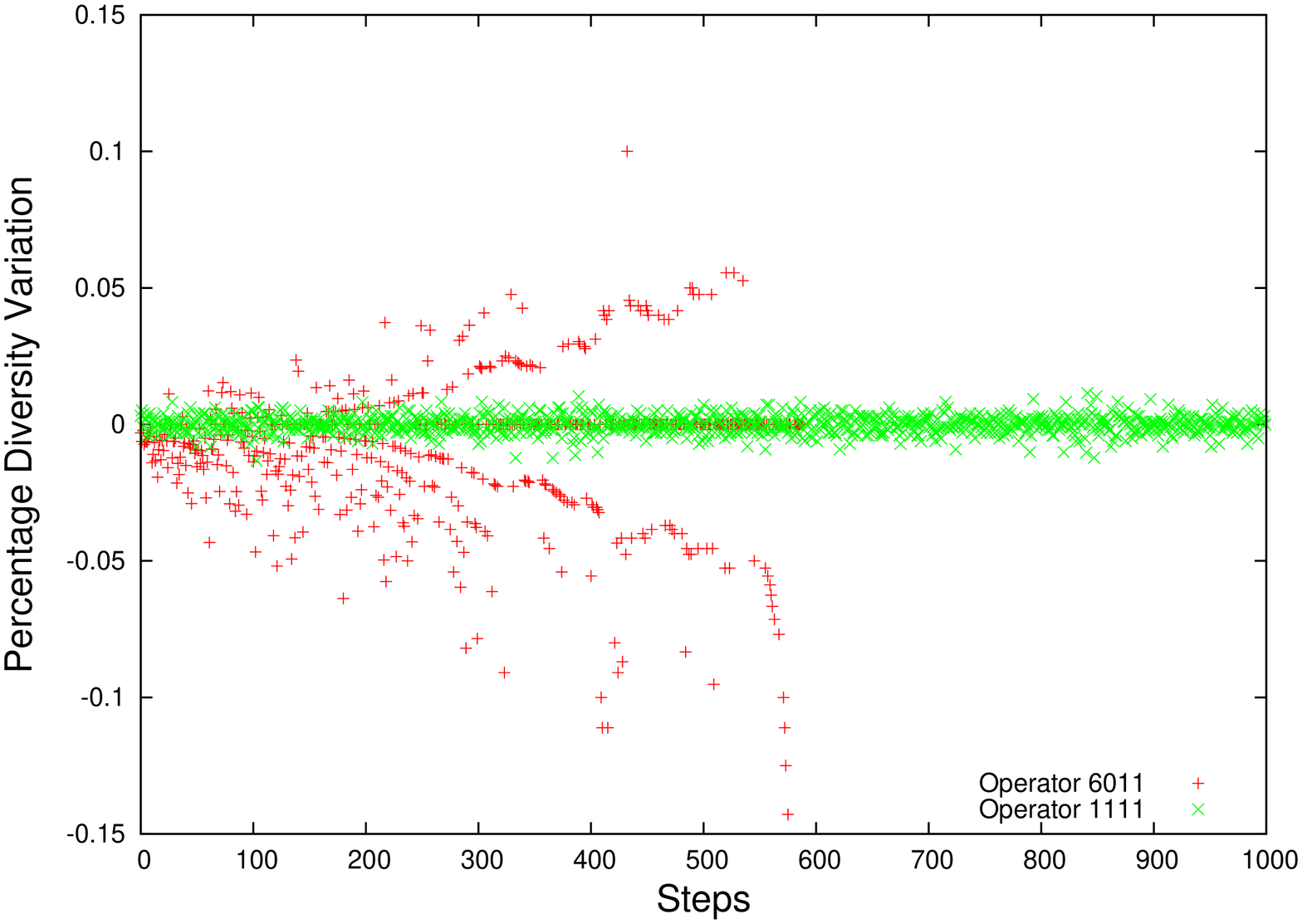}
(a) Impact on $\Delta D$, Operators $1111$ and $6011$.
\end{center}
\end{minipage} 
\end{center}
\caption{Assessing the impact of different operators over the variation of both quality and diversity. Instance $flat50-3$.}
\label{operatorsimpact}
\end{figure}

\begin{example}
Figure \ref{operatorsimpact} shows the impact on the two criteria observed on two runs of a EA capable of using just a single variation operator.
Figure \ref{operatorsimpact} (a) refers to the \comG{run based on the} application of an intensification oriented operator, labelled as $1111$; Figure \ref{operatorsimpact} (b) refers to the run based on the application of a
 a diversification oriented operator, labelled  as $6011$ (both will be defined in section \ref{Op}). Each plot shows the number of iterations performed by the EA on the x-axis and the percentage difference for the observed criterion (i.e., $\Delta Q$ on figure \ref{operatorsimpact} (a) and  $\Delta D$ on figure \ref{operatorsimpact} (b)). 
\end{example}

We can remark that the operator $6011$ has an impact over quality that is often bigger in magnitude
 than $1111$. Even though  this impact is often negative, leading to worsen the solution quality (figure \ref{operatorsimpact} (a)).
Accordingly, $1111$ has an unstable impact on diversity (figure \ref{operatorsimpact} (b)). Hence, when defining the 
experimental setting, the user has to keep on mind that the different criteria are often intertwined, and that choosing 
to favor a criteria does not mean that the other will be aimed to be constant during all the search process.

\paragraph{\textbf{Credit Assignment}}

Credit is defined as a measure that characterizes the reward obtained by an operator recently. In order to capture this \emph{typical} reward profile, the rewards assigned to operators are stored in a sliding window of size $T'win$. The Credit Assignment module works in the same way as the aggregated criteria computation method: it computes an aggregated credit for each operator, stored in a time window of size $T'win$ using a specific function $F'win$.
These credits are computed over a given period of time $T'win$ using a specific function $F'win$ which aggregates thus the successive rewards obtained by the operator. These values, that represent the operator's 
credit w.r.t. to their performances, represent the output of this component, and they are sent to the \emph{Operator Selection} component.  
Previous studies \citep{cec2009} have shown that  $T'win$ and $F'win$ have significantly less impact on the behavior of the controller than $Twin$ and $Fwin$. Therefore, to reduce the  combinatorial complexity of our analysis we will not address the issue of instantiating  these two parameters  and we will set them to the values used in that work.

\paragraph{\textbf{Operator Selection}}         

Once the credits have been computed for each operator, AOS must select one of them to recommend its application to the EA on the next iteration. This module determines the next operator to be applied by the  EA, according to  the credits (which are the input of the module). The operator selection is performed by means of a $Method$ which has to be defined by the user/developer. 
In this paper, after having performed preliminar experiences, we  use the probability matching selection rule (see Section \ref{previousworks}). This selection method is used with the $mean$ function for $Fwin$. We do not address the comparisons between methods, but such comparisons  can be found in \citep{cec2009,Fialho2010,Maturana2010,Tollo2011}. PM reduces the number of parameters in the selection method and  has shown good results on the problem we want to use for benchmarking (SAT problem).

\section{Experimental Setting}
\label{experimentalSetting}

This section describes the experimental setting used to explore the behavior of AOS. The EA is detailed in subsection \ref{bsea}, the operators in subsection \ref{Op}  and the benchmarks in subsection \ref{instances}.

\subsection{Basic Structure of the Evolutionary Algorithm}
\label{bsea}

Our purpose is to investigate how our controler influence the search process. To this aim we have chosen  to tackle the satisfiability 
problem (SAT)  \citep{Biere2009} for two  main reasons. On the one hand, many different problems can be encoded into SAT 
formalism, which provides different search landscapes and instances' structures for experiments. On 
the other hand, the EA we use is based on GASAT \citep{Lardeux2006}, that includes several variation operators 
whose performances are known according to previous studies \citep{Maturana2010}. 
The selection process consists of a  classic tournament over two randomly chosen 
individuals and the insertion process  replaces  the oldest individual of the population. 
The algorithm applies one operator at each step producing one individual from two parents. 

\begin{figure}[!ht]
\begin{center}
\includegraphics[width=0.75\textwidth]{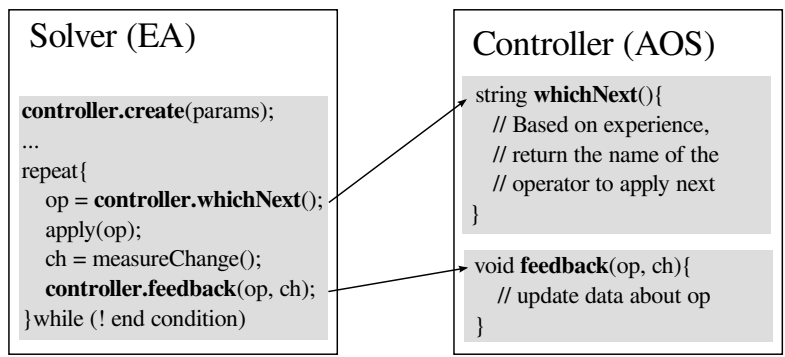}
\end{center}
\caption{Combination of EA and AOS}
\label{algoAOS}
\end{figure}

The combination of the Evolutionary Algorithm and the controller is sketched in figure \ref{algoAOS}.

Population size has been set to $30$. Since we want to observe the long term effects of the controller, the number of 
generations is set to $100\,000$ as default value. Concerning  computation time,  we stress out that, given the size of the individuals kept fixed, the execution time is constant for each operator application,  so the computation effort will be assessed by the number of crossovers performed during the search.  The EA and the controller have been coded in C++ and are available upon request.  Experiments have been run on a 280-core, 792 GFlop computer cluster.

\subsection{Operators definition}
\label{Op}

The operator to be applied is selected by the controller from a set of $20$ variation operators  (out of more than 300   operators defined by \citet{Maturana2010}).  These operators are specific to the SAT problem  and can defined by
 a combination of four basic features: 
\begin{enumerate}
\item selection of clauses that are false in both  
parents;
\item  action on each of the false clauses;
\item selection of clauses that are true in both parents;
\item action on each of the true clauses.
\end{enumerate}

\noindent An operator can be represented by a quadruplet $f_{1}f_{2}f_{3}f_{4}$, where  $f_{i}$ is the value  of feature $i$  according to table \ref{crossoverOps}. A variation operator is a function $S \times S \to S$, where $S$ is the set of all possible individuals (i.e., the search space).

\begin{table}[!ht]
\begin{center}
\begin{small}
\caption{Combinatorial definition of crossover operators\label{crossoverOps}}{
\begin{tabular}[t]{p{6cm}p{6cm}}
\hline \parbox{5cm}{\textbf{$i = 1$. Selection of clauses that are false in both parents}} & \parbox{5cm}{\textbf{$i = 2$. Action on each of the false clauses}} \\ 
\hline 
\parbox{5.5cm}{
 \begin{enumerate}\setcounter{enumi}{-1}
 \item do nothing
 \item select them in chronological order
 \item choose randomly one
 \item choose randomly one in the set of smallest clauses
 \item choose randomly one in the set of biggest clauses
 \item Fleurant Ferland \citep{fleurent96}\comJ{should we cite?}\comL{Done}
 \item flip variables which are identical in both parents 
 \end{enumerate}
}
&
\parbox{6cm}{
 \begin{enumerate}\setcounter{enumi}{-1}
 \item do nothing
 \item flip the variable that maximizes the number of true clauses and minimize the number of false clauses
 \item same as previous one, but the flip is not performed when the corresponding child's clause is already verified to be true
 \item flip all the variables
 \item flip the literal which appears less often in the others clauses
 \end{enumerate}
} \\
\hline
\parbox{5cm}{\textbf{$i = 3$. Selection of clauses that are true in both parents}} & \parbox{5cm}{\textbf{$i = 4$. Action on each of the true clauses}} \\ 
\hline 
\parbox{5.5cm}{
 \begin{enumerate}
 \item do nothing
 \item select them in chronological order
 \item choose randomly one
 \item choose randomly one in the set of smallest clauses
 \item choose randomly one in the set of biggest clauses
 \end{enumerate}
}
&
\parbox{6cm}{
 \begin{enumerate}
 \item do nothing
 \item set to true the variable that whose flip minimizes the number of false clauses
 \item set all the literals to true
 \item set to true the literal whose negation appears less often in the other clauses
 \item set all the literals to false
 \end{enumerate}
} \\
\hline  
\end{tabular}}
\end{small}
\end{center}
\end{table}

 \noindent All variables that remain unassigned in the resulting individual are valued using a classic uniform process$~$\citep{Sywerda1989}. 
In our experiments, we have selected the following operators, grouped according to their expected effect$~$\citep{Lardeux2006,Maturana2010}: 

\begin{itemize}
\item
\textbf{exploration:} $0011$, $0035$, $0015$, $4455$, $6011$; 
\item
\textbf{exploitation:} $1111$, 
$1122$, $5011$, $3332$, $1134$, $0022$, $2352$, $4454$, $1224$, $0013$; 
\item
\textbf{neutral:} $2455$, $4335$, $1125$, $5035$, $1335$. 
\end{itemize}
The following basic example  highlights how variation operators  may be used to get better individuals from a fitness point of view.

\begin{example}
\label{example}

Let us consider a  small SAT instance with three Boolean variables $a,b$ and $c$, and three 
 clauses $c_1 \equiv a \vee \neg b \vee \neg c$, $c_2 \equiv \neg a \vee  b$ and $c_3 \equiv \neg a \vee c$. The purpose of 
a SAT solver is to find a satisfying assignment, for instance $\{a \leftarrow 1, b\leftarrow 1, c\leftarrow 1\}$, where 
true and false are classically denoted as $1$ and $0$. 
In our EA, an individual (that represents a Boolean assignment) is a triple $(v_a, v_b, v_c)$, whose values represent the Boolean values assigned to $a$, $b$ and $c$. The fitness of an individual corresponds to the number of true clauses. The fitness of $(111)$ is thus $3$. The  operators are applied  on  two individuals in order to produce a new one. 

Let us consider the two assignments  $(1 1 0)$ and $(1 0 0)$ as input for an operator. 
$(1 1 0)$  satisfies  $c_1$ and $c_2$ but not $c_3$ and its fitness is $2$; $(1 0 0)$, whose  fitness is $1$,  satisfies only $c_1$. Therefore, $c_3$ is false for both assignments. If we consider the operator $1111$, it will select clause $c_3$ as common false clause and change variable $a$ to $0$ (since for $(1 1 0)$ it leads to $(0 1 0)$ and for $(1 0 0)$ it leads to $(0 0 0)$, both with fitness $3$). The resulting individual is obtained by setting $a$ to $0$, and finally, by completing uniformly: $c$ is   set to $0$, having the same value in $(1 1 0)$ and $(1 0 0)$; $b$ can be set either to  $1$ or to  $0$. In both cases, 
we get an individual  with  a fitness value $3$  (either $(0 1 0)$ or $(0 0 0)$), which improves the quality of the  population.

Notice that in this example we have considered a classic notion of \emph{fitness} function, which has to be maximised. However, given that SAT problem is often treated as a minimization problem (minimize the number of \emph{false} clauses), from here on we will use the term \emph{fitness} and \emph{quality} in terms of a minimization problem.

\end{example}

\subsection{Instances}
\label{instances}

In order to assess the general purpose of the controller, different representative SAT instances have been selected from the following  problems categories:
\begin{itemize}
 \item Random 3-SAT instances \citep{Cook1997};
 \item Random k-SAT instances sampled from the phase transition region \citep{hard}; 
 \item 3 Bit Colorable flat graphs \citep{tad};
 \item Subgraph Isomorphism Problems \citep{sgi};
 \item Hard Handmade instances \citep{simon}.
\end{itemize}
For more details, we forward the interested reader to the SAT competition's website \url{http://www.satcompetition.org/}.
The main instance's features  are reported in Table \ref{bench}. For each experiment on the same instance, the same 
initial population is used.

\begin{table}[!ht]
   \begin{center}
    \caption{Benchmark features}
\begin{small}
\begin{tabular}{llrr}

ID& Instance name & variables &clauses\\

\multicolumn{4}{c}{Random 3-SAT}\\
\hline

1&F500 &500& 2150\\

\multicolumn{4}{c}{Random k-SAT instances}\\
\hline

2&unif-k7-r89-v65-c5785-S1481196126 &65 &5785\\
3&unif-k7-r89-v65-c5785-S1678989107 &65 &5785\\
4&unif-k7-r89-v65-c5785-S2099893633 &65 &5785\\
5&unif-k7-r89-v65-c5785-S316555917 &65 &5785\\
6&unif-k7-r89-v65-c5785-S461794864 &65 &5785\\
7&unif-k7-r89-v75-c6675-S1299158672 &75 &6675\\
8&unif-k7-r89-v75-c6675-S1534329206 &75 &6675\\
9&unif-k7-r89-v75-c6675-S1572638390 &75 &6675\\
10&unif-k7-r89-v75-c6675-S1785258608 &75 &6675\\

\multicolumn{4}{c}{3 Bit Colorable}\\
\hline

11&flat50-293 &150& 545\\
12&flat50-297 &150& 545\\
13&flat50-298 &150 &545\\
14&flat50-299 &150 &545\\
15&flat50-3 & 150&545\\
16&flat50-30 &150 &545\\

\multicolumn{4}{c}{Subgraph Isomorphism Problems}\\
\hline

17&new-difficult-20 &360& 15466\\
18&new-difficult-21  &399& 18184\\
19&new-difficult-22 &440& 22276\\
20&new-difficult-23 &483 &25396\\
21&new-difficult-24 &528 &30728\\
22&new-difficult-26 &624 &38944\\
23&new-difficult-28 &728 &48442\\
24&satsgi-n23himBHm26 &598 &14076\\
25&satsgi-n23himBHm27 &621 &14927\\
26&satsgi-n25himBHm27 &675 &16900\\
27&satsgi-n25himBHm29 &725 &18875\\
28&satsgi-n28himBHm30 &840 &23548\\
29&sgi-difficult4 &483 &15156\\

\multicolumn{4}{c}{Hard handmade}\\
\hline

30&sgi-difficult7 &728 &28986\\
31&Simon &2424& 14812\\
32&3bit &8432&31310\\

\end{tabular}
\end{small}
\label{bench}
 \end{center}
 \end{table}

\vspace{3ex}

\section{Operators Management}
\label{opman}

The goal of the controller is to manage the trade-off between exploitation and exploration. Therefore, the operators design and choice are of the utmost importance, since some operators lead to the concentration of the population into specific areas of the search space (exploitation), while other operators may be more related to exploration, depending on the current search state.

In this section we study some relevant features of our controller. Operators management in presence of null operators is discussed in section \ref{nul}, while the definition of search strategies is presented in section \ref{var}.

\subsection{Experiments with null operators}\label{nul}

We use the term {\em null operator} to identify operators that take two individuals as input, and  outputs one of them, thus having no effect over the population if used jointly with an appropriate insertion process (they replace  an individual by itself, and therefore the variation of the criteria is $0$). 
%

\begin{figure}[!ht]
\begin{center}
\begin{minipage}[t]{0.45\textwidth}
\begin{center}
\includegraphics[width=\textwidth]{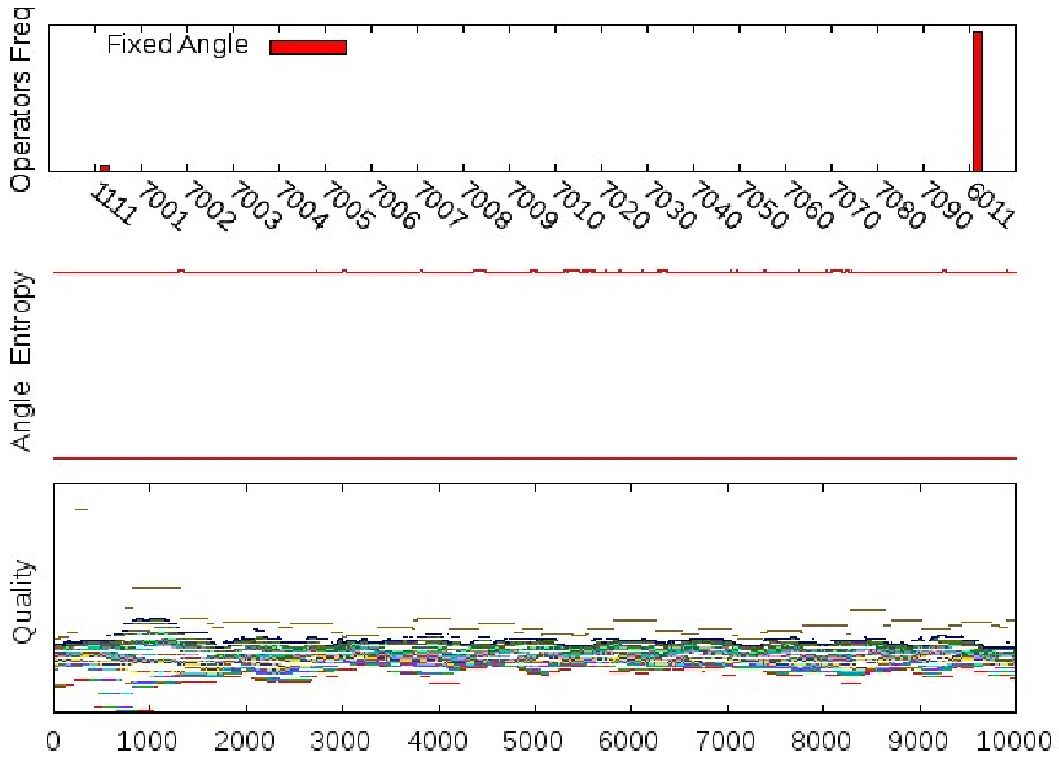}
(a)  $\theta = 0$
\end{center}
\end{minipage}
\begin{minipage}[t]{0.45\textwidth}
\begin{center}
\includegraphics[width=\textwidth]{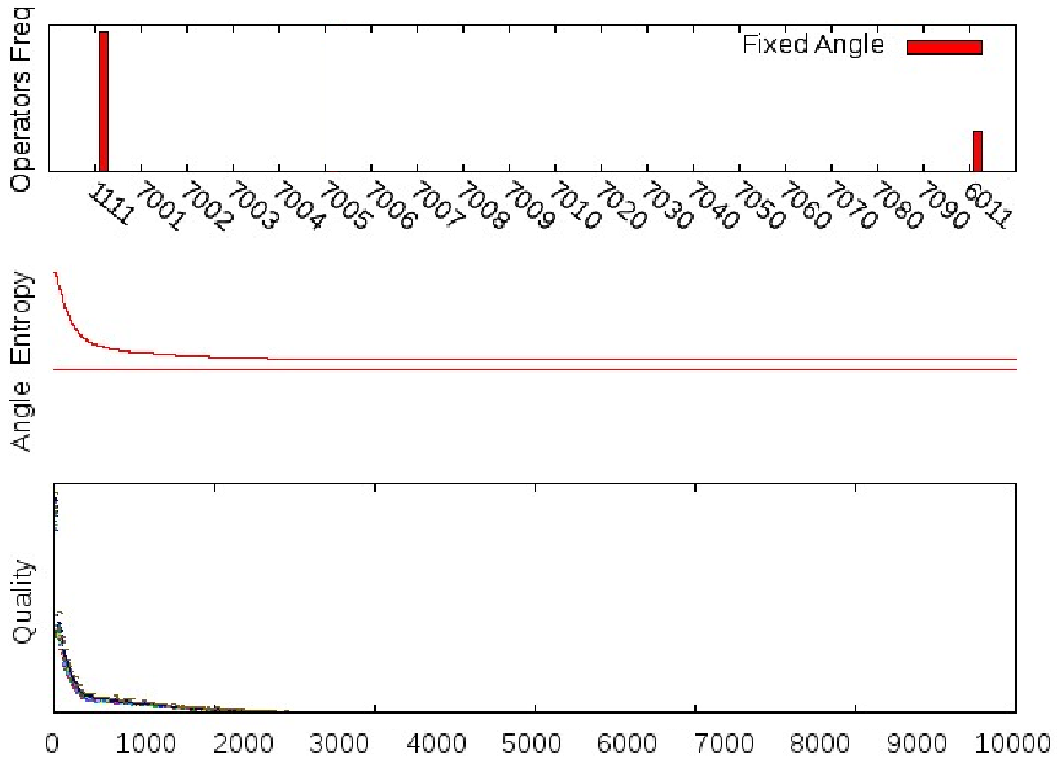}
(b) $\theta = \pi / 2$  
\end{center}
\end{minipage}
\end{center}
\caption{Experiment with null operators, different fixed $\theta$ values\label{neutfix}.  Instance 3bits.}
\end{figure}

\begin{figure}[!ht]
\begin{center}
\includegraphics[width=6cm]{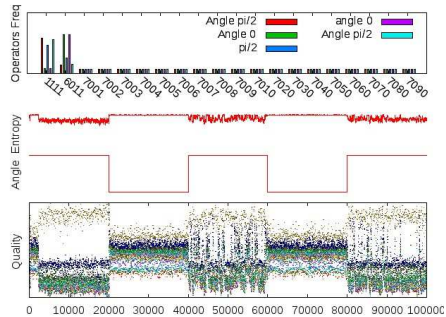}
\caption{Experiment with null operator, changing the angle\label{neutchange}. Instance Simon.}
\end{center}
\end{figure}

We have carried out experiments using a set of operators containing an exploration-oriented operator ($6011$), an exploitation-oriented one ($1111$) and  $18$ null operators (identified by the tuples $70**$ in figures \ref{neutfix} and  \ref{neutchange}). Our purpose is to check whether  the controller discriminates amongst the proposed operators according to the desired level of exploitation-exploration.  In the following pictures, we show in top part the frequency of application of all operators (labelled on the x-axis). The remaining three parts show, respectively, the variation of entropy, the variation of the $\theta$ parameter (labelled as \emph{angle}) and the   fitness evolution of all individuals over time (steps are labelled in the x-axis). \comG{I will insert a comment about the y axis, have to find what I have written in an old paper}

The controller is expected to identify the null operators. In the same time, the controller must apply the non-null operator that fits the  required behavior (defined by $\theta$). 
Notice that null operators are not significantly used, and the proportion of application on non-null operators produced the expected effect on the search.



By defining a sequence of changes of policy throughout the search, we can define a search \emph{strategy} (see section \ref{sec:varpolicy}). This is done by varying the value of \emph{SDir} in the reward computation module, i.e., the angle $\theta$. Figure \ref{neutchange} shows the application frequency and the behavior (in terms of entropy and quality) when alternating between extreme angles.


We can notice that the controller also succeeds in detecting the suitable operators according to the different required search direction, and relegate null operators to a second place. \comG{I think we'd better use all figures from PM. }\comS{??}\comL{For a better understanding, it would be better not to dwell on PM and MAB} \comG{Sure, but we are using pictures drawn from MAB experiments, whilst in the introduction we say that we will be using PM, so I will change the pictures accordingly.}

\subsection{Search strategies}\label{var}
\label{sec:varpolicy}

As stated in the introduction, we are  interested in considering dynamic  policies during the search. This  defines either a predefined or a dynamic change between policies that  allow us to guide the search according to a previously defined or a reactive schedule, respectively. In this work we explore the following simple strategies that guide the search by changing the value of the angle $\theta$:

\begin{itemize}
\item \emph{INCREASE}: To split the execution time into several epochs and to increase the angle value in equally distributed levels in $[0,\frac{\pi}{2}]$.
\item \emph{DECREASE}: To split the execution time into several epochs and to decrease the angle value in equally distributed levels in $[0,\frac{\pi}{2}]$.
\item \emph{ALWAYSMOVING}: To split the execution time into several epochs and to alternate the angle value between $0$ and $\frac{\pi}{2}$ (as shown in the previous section).
\item \emph{REACTIVEMOVING\comL{DYNAMICALWAYSMOVING}}: Similar to \emph{ALWAYSMOVING} but setting $\theta$ to $\frac{\pi}{2}$ when the entropy value is less than $0.9$ and to $0$ when the quality has not increased for $200$ consecutive iterations.
\end{itemize}

\begin{figure}[th]
\begin{center}
\begin{minipage}[t]{0.4\textwidth}
\begin{center}
\includegraphics[width=\textwidth]{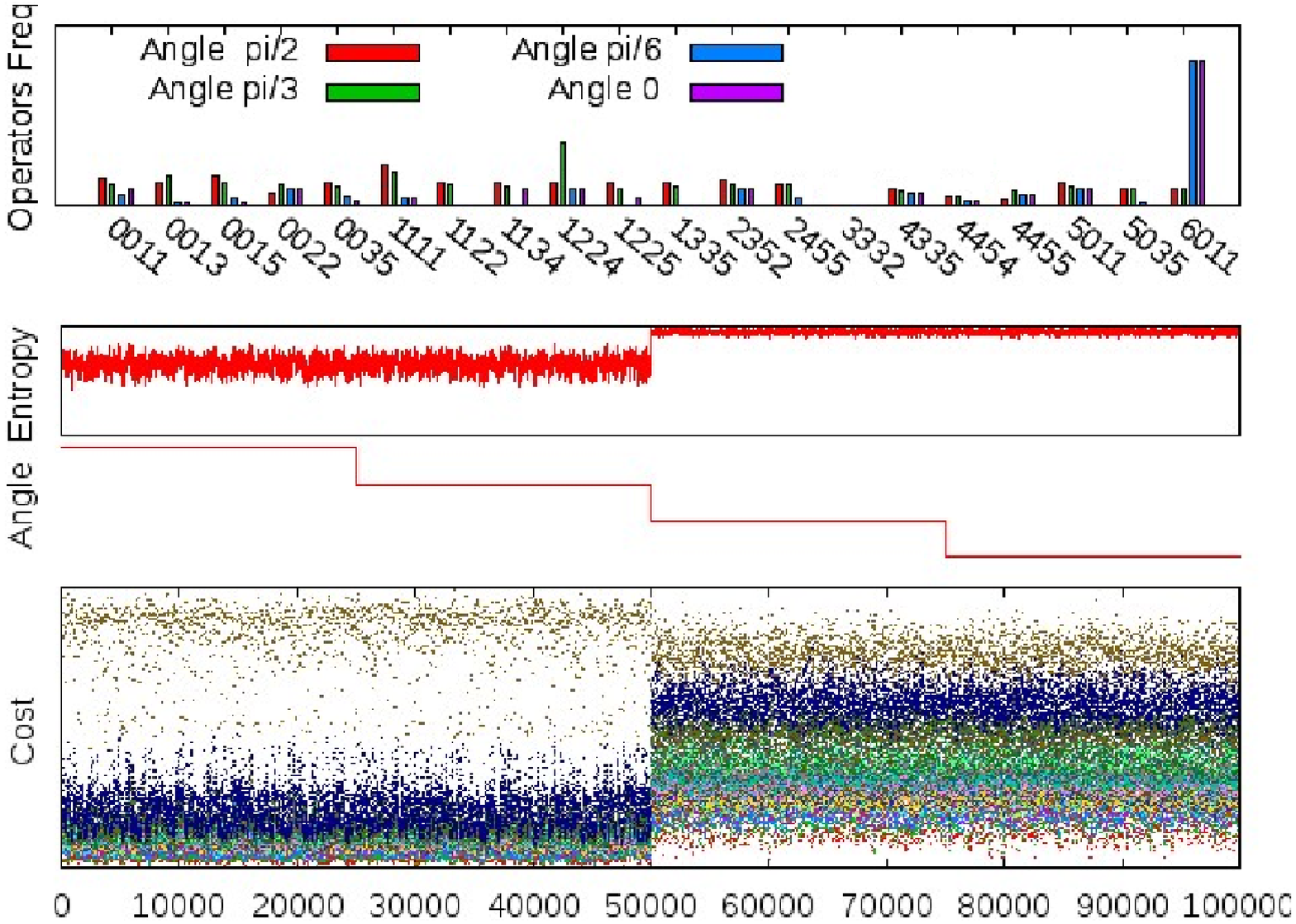}
DECREASE strategy
\end{center}
\end{minipage}
\begin{minipage}[t]{0.4\textwidth}
\begin{center}
\includegraphics[width=\textwidth]{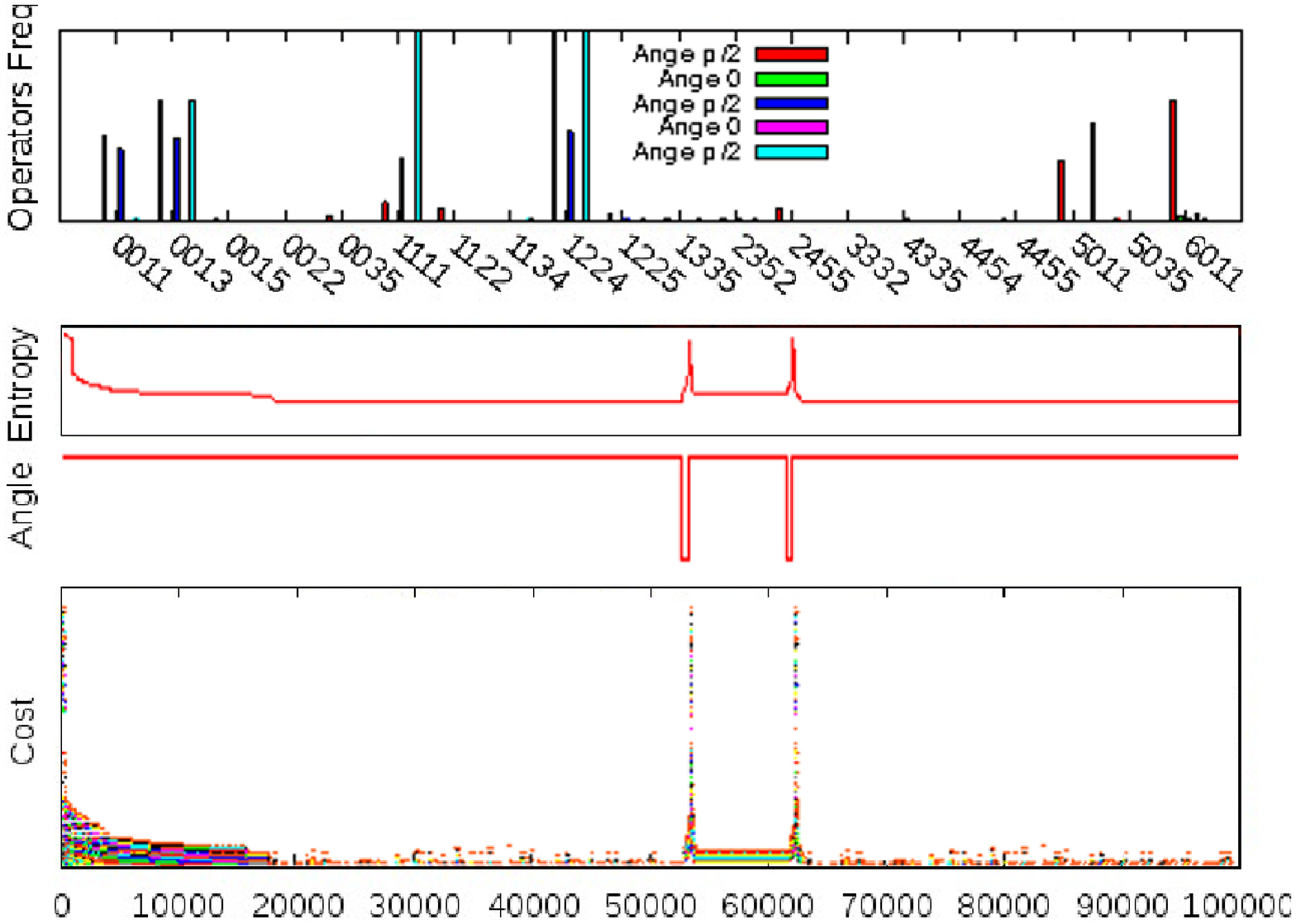}
REACTIVEMOVING strategy \comJ{Since AM was already shown in the previous picture, it would be better to include RM\comL{AM-DYN} here} \comG{Yes I can do it}
\end{center}
\end{minipage}
\caption{Dynamic strategies. Instance Simon}
\label{alwaysoperators}
\end{center}
\end{figure}

\noindent In order to show how AOS orient the search by changing the angle, figure \ref{alwaysoperators} presents the variation of the population's  mean quality and diversity when the  value of angle $\theta$ changes in the range $[0,\frac{\pi}{2}]$ for two different strategies, using the operators listed in section \ref{Op}.

We remark that the controller succeeds in determining, for each epoch corresponding to a given $\theta$
value, which operator has to be used in order to foster the given policy (operator $6011$ for the exploration epochs and $1111$ for the exploitation epochs). \comG{JORGE: I have put the comment you asked about picture 6 here. But are you sure that this is the good place? Furthermore, I have put the reactive moving picture. But it is really akward, since the exploration phases are really small. I have to put some more comments. Do you agree?}

\section{Solving Performance}
\label{sec:performances}

In this section we study the effect of the controller in terms of improvement of the solutions obtained by the EA. In section \ref{randomparamils}, we show that the introduction of the
controller leads to solutions whose quality is comparable (when not better) with regards to other selection methods. In section \ref{robust} we study the behavior of the diverse dynamic strategies presented in section \ref{var}. In section \ref{memetic} we discuss  results obtained by adding a tabu search mechanism to the GA+controller, in order to escape from local optima and to get  better performances.

\subsection{Controller vs. Tuning Methods}
 \label{randomparamils}

We start our analysis by comparing our combination EA+controller with two other solving approaches: 
\begin{itemize}
\item
an EA that uses a uniform random selection of the operators introduced in section \ref{opman}; 
\item
an EA whose  operator application rates have  been optimally tuned by using ParamILS (see section \ref{previousworks}).\footnote{As for ParamILS implementation, we have defined a  
  discrete set of values for the $20$ parameters, 
  consisting of $11$ equi-distanced possible values in the range $[0~1]$. 
  We have used the Focused-ILS variant, setting the cut-off time to $70$, since for at least $75 \%$ of 
  the instances, GASAT completes within $70$ seconds. The overall time budget allocated for the whole process 
  has been set to $20\,000$ seconds \citep{Hoos2012}.} The operator selection is achieved according to a roulette-wheel mechanism whose operators' application probability are known \textit{a priori}.
\end{itemize}

The controller is first used with fixed search policies  $\theta = \frac{\pi}{4}$, $\theta = \frac{\pi}{2}$. Note that a fixed policy does not mean that the application rates of the operators are fixed but rather thah they are adapted dynamically in order to maintain the desired fixed trade-off between quality and diversity. 

Results  are presented in table \ref{resrandom}, where the  quality of the  best solution found over $30$ runs is reported  for the different strategies  labelled on columns.\footnote{Column ID represents the instance number, see table \ref{bench}.} 
The best solution amongst strategies are boldfaced.

 \begin{table}[!ht]
  \begin{center}
\caption{Best solution fitness for Controller ($\theta = \frac{\pi}{4}$ and $\theta = \frac{\pi}{4}$), Random Selection and ParamILS.}{
\begin{footnotesize}
\begin{tabular}{l |r r r r r r r r}
ID.	&\multicolumn{2}{c}{$\theta = \frac{\pi}{4}$}	&\multicolumn{2}{c}{$\theta = \frac{\pi}{2}$} &\multicolumn{2}{c}{Random}	&\multicolumn{2}{c}{ParamILS}\\	
&Min&Std&Min&Std&Min&Std&Min&Std\\		

\hline
\\
&	\multicolumn{8}{c}{Random 3-SAT}\\
\hline
1	&5	&13.09	&\textbf{1} &11.7	& 53		&6.72&59&5.06	\\
\\
&	\multicolumn{8}{c}{Random k-SAT instances}\\ 
\hline
2	&12	&16.41	& 12	&1.35&\textbf{11}&1.58		&14		&1.56\\
3	&15	&16.62	& \textbf{12}	&1.69&14&1.99		&13	&1.55\\
4	&14	&16.65	& \textbf{2}	&3.19&13&1.47		&9	&	2.57\\
5	&12	&15.96	& \textbf{7}   &2.56& 12&1.79	&\textbf{12}&1.70\\
6	&15	&15.31	& \textbf{2} 	&3.23& 15&	1.71	&\textbf{12}	&1.90\\
7	&17	&20.68	& 15	&1.80& 17&	1.66	&\textbf{14}		&2.47\\
8	&19	&19.41	&\textbf{2} 	&3.80& 16 &1.72&15	&1.87\\
9	&17	&19.82	&\textbf{5} 	&3.39& 17	&2.08	&17	&	2.09\\
10	&14	&20.27	&\textbf{4}	&2.43& 17	&2.04	&17	&	1.36\\
\\
&	\multicolumn{8}{c}{3 Bit Colorable}\\
\hline
11	&1&1.27	&\bf{0}&2.04	&13&1.90		&9	&1.60	\\
12	&13&1.96		&\bf{0}&1.23	&12	&1.62	&11	&0.75	\\
13	&13	&2.03	&\bf{0}& 2.16  &11	&1.61	&10		&1.47\\
14	&3&	1.17	&\bf{0}&1.65	&12	&1.58	&9		&1.26\\
15	&14	&1.68	&\bf{0}&2.17	&12	&1.65	&8	&1.90	\\
16	&4&	9.72	&\bf{0}&2.61	&11&	0.66	&\textbf{0}	&1.18\\
\\
&	\multicolumn{8}{c}{Subgraph Isomorphism Problems}\\ 
\hline
17		&\textbf{3}&11.35	&4&4.33		&11	&0.86	&9	&1.92	\\
18		&\textbf{3}&15.79	&4&	3.38	&9	&0.84	&8	&	1.04\\
19		&\textbf{5}&2.37	&\textbf{5}&	1.71	&9	&1.08	&10	&	0.70\\
20		&\textbf{3}&3.68	&5&	1.85	&11	&0.76	&11	&	1.00\\
21		&\textbf{3}&4.31	&\textbf{3}&	2.06	&13	&0.88	&11	&	2.72\\
22		&\textbf{13}&5.44	&5&	3.02	&14	&0.99	&\textbf{13}&1.25	\\
23		&4	&4.89		&8&	3.42	&\textbf{3}	&0.77	&\textbf{3}&	1.58\\
24	      &\textbf{0}	&5.65&\textbf{0}&3.49		&2&1.02		&5&	0.98	\\
25		&\textbf{0}&6.34	&\textbf{0}&3.17		&6&0.99		&2&2.07		\\
26		&\textbf{0}&5.65	&\textbf{0}&	3.96	&6&	1.44	&7&	1.27	\\
27		&\textbf{0}&7.96	&\textbf{0}&	3.94	&5&	1.30	&5&	1.47	\\
28		&\textbf{0}&6.72	&\textbf{0}&	5.10	&5&	1.12	&7&	1.02	\\
29		&\textbf{0}&8.65	&\textbf{0}&	3.96	&7&	1.58	&6&	1.43	\\
30		&1&10	&\textbf{0}&	5.69	&8&	4.96	&6&	1.25	\\
\\
&	\multicolumn{8}{c}{Hard handmade}\\
\hline
31		&\textbf{19}&27.26	&20	&15.39&70	&	33.19		&59	&	7.79\\
32		&68	&100.48	&\textbf{21}&65.53	&157	&		76.37	&138&	18.66	\\
\end{tabular}
\end{footnotesize}}
\label{resrandom}
 \end{center}
 \end{table}

\comJ{I found this paragraph either contradictory or difficult to understand (j'attends)}


We remark that the $\theta = \frac{\pi}{4}$ controller provides better results than the random selection. 
As for   ParamILS, we remark that  it shows good performances, especially 
when tackling Random k-SAT instances. Anyhow, when tackling  these instances, ParamILS results are not 
significantly different from the fixed angle's ones. On other instance instead, the controller (fixed angles or Alwaysmoving) 
performs always better than ParamILS.

We also remark that just focusing on quality ($\theta =  \frac{\pi}{2}$) represents actually a good stand-alone criterion for some instances, but nevertheless fails to reach good solutions for many instances.

Indeeed, the choice of the suitable operators with regards to a given compromise between  criteria has to be coupled by a strategy that determines how much time has to be spent in achieving this given compromise. If the population recent history indicates that  no further improvements can be reached with regards to this compromise, keeping on  having the same controller setting can result in a waste of computational time, which could be more effectively used otherwise. The fixed policies leads to results that are not satisfactory since assignments are hardly found for some instances and we have to turn to more dynamic control strategies.

\subsection{Experiments with dynamic strategies}
\label{robust}
In order to improve results obtained in section  \ref{randomparamils},  we are interested in using the strategies described in section \ref{sec:varpolicy}. In particular, we will use the dynamic strategy labelled as  \emph{REACTIVEMOVING}\comL{\emph{Instead of DYNAMICALWAYSMOVING}},  in which $\theta$ values switches between $0$ and $\frac{\pi}{2}$ according to the state of the search. 

In order to assess the performance of these strategies, we use a fixed angle policy ($\theta=\frac{\pi}{2}$) as baseline and a steady-state GA  \citep{Lardeux2006} that  uses the optimized operator  CC ($1111$ w.r.t. our operator taxonomy).  Note that this crossover has been optimized using time consuming experiments on several SAT instances. Table \ref{mstd} shows results obtained by the the diverse strategies labelled on columns.

\begin{table}[!ht]
  \begin{center}
\caption{Best and standard deviation of fitness for several Controller settings.}{
\begin{footnotesize}
\begin{tabular}{r|   r r r r r r r r r r r r}
ID&						\multicolumn{2}{c}{$\frac{\pi}{2}$ }			&\multicolumn{2}{c}{INC}			&\multicolumn{2}{c}{DEC}			&\multicolumn{2}{c}{AM} &\multicolumn{2}{c}{RM} &\multicolumn{2}{c}{CC ONLY}\\
\hline 
&			Min	&Std		&Min	&Std		&Min	&Std		&Min	&Std &Min	&Std &Min	&Std\\		
&	\multicolumn{10}{c}{Random 3-SAT}\\
\hline
1			&53	&	6.72		&5	&13.09		&\textbf{1}	&11.70		&5	&6.18		&2		&11.24		&7		&5.43\\
\\
&	\multicolumn{10}{c}{Random k-SAT instances}\\ 
\hline
2			 		&12	&1.35		&\textbf{0}	&1.41		&1		&1.15		&6		&1.09	& 1 		& 2.84 & 1 & 1.15\\
3			 	&12	&1.69		&2		&0.66		&\textbf{1}	&0.57		&11		&1.35	& \textbf{1} 	& 1.55 & 1 & 0.9\\
4			 	&2	&3.19		&\textbf{0}	&0.61		&\textbf{0}	&0.55		&4		&1.19	&\textbf{0}	&4.4					&1&1.11\\
5			 	&7	&2.56		&\textbf{0}	&1.08		&\textbf{0}	&1.3		&12		&3.18	&\textbf{0}	&1.76					&0&1.44\\
6			 		&2	&3.23		&1		&0.001		&1		&0.1		&12		&2.02	&2		&1.48				&1&1.1\\
7			 	&15	&1.80		&1		&1.45		&2		&1.1		&1		&1.64	&\textbf{0}	&0.58					&2&1.02\\
8			 	&2	&3.80		&1		&0.001		&1		&0.1		&17		&0.88	&\textbf{0}	&1.75					&2&0.95\\
9			 	&5	&3.39		&1		&0.65		&1		&0.81		&3		&1.31	&\textbf{0}	&4.53					&1&1.05\\
10			 		&4	&2.43		&\textbf{0}	&0.3		&\textbf{0}	&0.24		&17		&1.96	&1		&4.1				&1&1.03\\
 \\
&	\multicolumn{10}{c}{3 Bit Colorable}\\
\hline
11			 	&\textbf{0}	&2.04		&2		&0.70		&1	&0.67		&4	&1.52	&1	&4.4							&2&1.53\\
12			 	&\textbf{0}	&1.23		&1		&0.63		&1	&0.51		&5	&4.09	&1	&1.62							&2&1.15\\
13			 	&\textbf{0}	&2.16		&\textbf{0}	&0.81		&1	&0.8		&4	&21.09	&1	&1.71							&1&1.7\\	
14			 	&\textbf{0}	&1.65		&1		&0.84		&1	&0.53		&4	&1.42	&1	&1.35							&1&1.63\\
15			 	&\textbf{0}	&2.17		&1		&0.001		&1	&0.001		&7	&2.10	&\textbf{0}	&4.7						&1&1.47\\
16			 	&\textbf{0}	&2.61		&2		&0.001		&1	&0.001		&11	&1.43	&0	&2.4							&1&1.78\\
\\
&	\multicolumn{10}{c}{Subgraph Isomorphism Problems}\\ 
\hline
17			 	&4	&4.33		&2	&0.001		&3	&0.001		&12	&1.27	&\textbf{0}	&0			&0&0.38\\
18			 	&4	&3.38		&3	&1.25		&2	&1.33		&16	&1.27	&\textbf{0}	&0			&0&0.54\\
19			 	&5	&1.71		&2	&0.47		&3	&0.48		&13	&0.84	&\textbf{0}	&0&0&0.9\\
20			 	&5	&1.85		&2	&0.3		&2	&0.001		&13	&1.55	&\textbf{0}	&0&0&0\\
21			 	&3	&2.06		&2	&0.001		&3	&0.001		&3	&1.52	&\textbf{0}	&0&0&0.25\\
22			 	&5	&3.02		&2	&0.001		&3	&0.001		&16	&4.11	&\textbf{0}	&0&0&0.34\\
23			 	&8	&3.42		&3	&0.71		&3	&0.65		&3	&0.84	&\textbf{0}	&0&0&0.04\\
24			 	&\textbf{0}	&3.49		&\textbf{0}	&0.44		&\textbf{0}	&0.53		&6	&0.92	&\textbf{0}	&0					&0&0\\
25			 	&\textbf{0}	&3.17		&\textbf{0}	&0.47		&\textbf{0}	&0.57		&8	&1.51	&\textbf{0}	&0					&0&0\\
26			 		&\textbf{0}	&3.96		&\textbf{0}	&0.4		&\textbf{0}	&0.39		&9	&1.90	&\textbf{0}	&0&0&0\\
27			 	&\textbf{0}	&3.94		&\textbf{0}	&0.85		&\textbf{0}	&1.28		&9	&1.31	&\textbf{0}	&0&0&0.45\\
28			 	&\textbf{0}	&5.10		&\textbf{0}	&0.3		&\textbf{0}	&0.63		&17	&3.85	&\textbf{0}	&0&0&0.6\\
29			 	&\textbf{0}	&3.96		&\textbf{0}	&0.001		&\textbf{0}	&0.001		&9	&1.64	&\textbf{0}	&0.18&0&0.54\\
30			 &\textbf{0}	&5.69		&\textbf{0}	&0.97		&\textbf{0}	&0.90		&3	&0.74	&\textbf{0}	&0.37&0&0.3\\
\\
&	\multicolumn{10}{c}{Hard handmade}\\
\hline
31			 &20	&15.39		&\textbf{9}	&12.26		&10	&19.36		&96	&19.07	&17	&9.4				&21&4.48\\
32			 &21	&65.53		&53	&0.64		&51	&0.54		&12	&2.00	&\textbf{9} &243.22				&15&7.68\\
\end{tabular}
\end{footnotesize}}
\label{mstd}
 \end{center}
 \end{table}

The policies that perform ``blind" $\theta$ variations (INCREASE, DECREASE and AM) can offer better results since they provide the mechanism to  escape from the current search local optimum, but improvements are hindered by  the  inefficient use of the computation time. The ReactiveMoving\comL{instead of DynamicAlwaysMoving} instead, offers the best results, given its capability to adapt to the search scenario.\footnote{  We have also tried to implement a dynamic version of Increase and Decrease, but in these approaches  we have faced the problem to implement the idle mechanism w.r.t. intermediate angle values: it is not clear when the value of $\theta$ has  to be changed  when its value is different from  $0$ or $ \frac{\pi}{2}$. This investigation is left for further works.}


 We want to remark that the ReactiveMoving strategy offers results which are comparable to the ones obtained by the CC-based algorithms. Please notice that the CC-based algorithm has been tuned by means of time-consuming experience, whilst ReactiveMoving do not require preliminary experiments.

\subsection{Memetic Algorithms and different operators set}
\label{memetic}

In memetic algorithms \citep{moscato}, the solution generated by variation operators - typically crossover or recombination operators - are refined by a local search algorithm.

\begin{figure}[!ht]
\begin{center}
\includegraphics[width=0.75\textwidth]{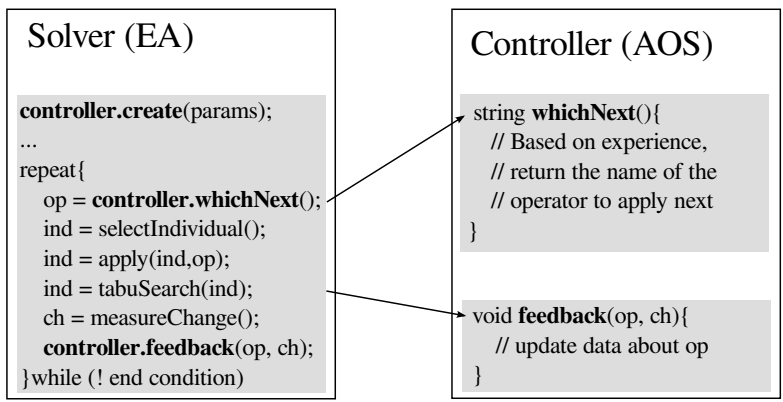}
\end{center}
\caption{Combination of EA+Tabu and AOS}
\label{algoTabu}
\end{figure}

The integration of a Tabu Search\citep{Glover1999} mechanism in an EA for the SAT problem has been proposed in \citep{Lardeux2006}, showing that this combination leads to improvement  of the initial performances. 
This memetic algorithm is sketched in figure \ref{algoTabu}. 
\%comG{no need for including the function, we are never using it}

The Tabu parameters have been set after preliminary experiences: lenght of the tabu list is $10$\% of the number of variables in the instance, and stopping criterion is either a number of variable flips or when a satisfiable solution has been obtained \comG{I would explain a bit better}. We have conducted experiments by adding the Tabu mechanism to the following strategies:

\begin{itemize}
\item Increasing;
\item ReactiveMoving\comL{Instead of AlwaysmovingDynamic};
\item Fixed Angle ($\theta= \frac{\pi}{2}$);
\item One operator only, no control and no strategy (as for section  \ref{robust}, the operator is the CC ($1111$) 
, i.e.,  the best performing exploitation operator).
\end{itemize}

\noindent We have compared them with a stand-alone Tabu Search 
\footnote{Length of the Tabu list has been set to $= 10$ percent of the number of variables. The process ends when $1\,0000\,000$ iterations have been performed.} and with the steady-state  GA based on the CC operator, which is still nowadays a reference EA for SAT. The results of this investigation are summarised in table \ref{tabuandsingleop}.

\begin{table}[!ht]
  \begin{center}
\caption{Best and standard deviation of fitness for several Controller settings.\label{tabuandsingleop}}{
\begin{footnotesize}
\begin{tabular}{r|   r r r r r r r r r r}
ID						&\multicolumn{2}{c}{TABU + INC}			&\multicolumn{2}{c}{TABU + RM}	 &\multicolumn{2}{c}{TABU + $ \theta = \frac{\pi}{2}$} &\multicolumn{2}{c}{TABU + CC} &\multicolumn{2}{c}{TABU ONLY}  \\
\hline 
&			Min	&Std		&Min	&Std		&Min	&Std		&Min	&Std &Min	&Std\\		
&	\multicolumn{10}{c}{Random 3-SAT}\\
\hline
1			&0 	&0,5 		&0	&0.5	 	&0	&0,84		&0	&0,75		&9	&1,17\\

\\
&	\multicolumn{10}{c}{Random k-SAT instances}\\ 
\hline
2			&0	&0.4		&0	&0.42		&0	&0.42		&0	&0,7		&5	&0.79\\
3		 	&0	&0.67		&0 	&0.34		&0 	&0.58		&0	&0.77		&5	&0.99\\
4 			&0	&0.37		&0	&0		&0 	&0.51		&0 	&0.47		&5	&0.87\\
5			&0 	&0.25		&0 	&0.39		&0 	&0.4		&0 	&0.67		&5	&0.89\\
6			&0	&0.37		&0	&0.38		&0	&0.38		&0	&0.68		&4 	&0.78\\ 
7			&0 	&0.34		&0 	&0.25		&0 	&0.34		&0 	&0.26		&6	&0.99\\
8			&0 	&0.41		&0 	&0.44		&0 	&0.36		&0 	&1.2		&5	&0.7\\
9			&0 	&0.18		&0 	&0.21		&0 	&0.18		&0 	&0.54		&5	&0.7\\
10			&0 	&0.31		&0 	&0.41		&0 	&0.31		&1 	&0.55		&6	&0.73\\
 \\
&	\multicolumn{10}{c}{3 Bit Colorable}\\
\hline
11			&0	&0		&0	&0		&0	&0		&0	&0		&0	&0\\
12			&0	&0		&0	&0		&0	&0		&0	&0		&0	&0\\				 	
13			&0	&0		&0	&0		&0	&0		&0	&0		&0	&0	\\	
14			&0	&0		&0	&0		&0	&0		&0	&0		&0	&0			 		\\
15			&0	&0		&0	&0		&0	&0		&0	&0		&0	&0			 		\\
16			&0	&0		&0	&0		&0	&0		&0	&0		&0	&0			 		\\
\\
&	\multicolumn{10}{c}{Subgraph Isomorphism Problems}\\ 
\hline
17				&0	&0		&0	&0		&0	&0		&0	&0		&0	&0		 	\\
18				&0	&0		&0	&0		&0	&0		&0	&0		&0	&0		 	\\
19				&0	&0		&0	&0		&0	&0		&0	&0		&0	&0		 	\\
20				&0	&0		&0	&0		&0	&0		&0	&0		&0	&0		 	\\
21				&0	&0		&0	&0		&0	&0		&0	&0		&0	&0		 	\\
22				&0	&0		&0	&0		&0	&0		&0	&0		&0	&0		 	\\
23				&0	&0		&0	&0		&0	&0		&0	&0		&0	&0		 	\\
24			&0	&0		&0	&0		&0	&0		&0	&0		&0	&0			 	\\
25			&0	&0		&0	&0		&0	&0		&0	&0		&0	&0			 	\\
26			&0	&0		&0	&0		&0	&0		&0	&0		&0	&0			 	\\
27			&0	&0		&0	&0		&0	&0		&0	&0		&0	&0			 	\\
28			&0	&0		&0	&0		&0	&0		&0	&0		&0	&0			 	\\
29			&0	&0		&0	&0		&0	&0		&0	&0		&0	&0			 	\\
30			&0	&0		&0	&0		&0	&0		&0	&0		&0	&0			 	\\
\\
&	\multicolumn{10}{c}{Hard handmade}\\
\hline
31			&12	&3.47		&15 	&2.61		&16 	&2.98		&21 	&3.02		&22	&1.64	\\
32			&3	&1.5		&4 	&1.2		&3	&1.41		&4 	&1.27		&15	&2.84	\\
\end{tabular}
\end{footnotesize}}
 \end{center}
 \end{table}

 By comparing the results with those outlined in table \ref{mstd}, we can observe that by adding a simple Tabu Search, the performance of the   
controlled GA is better 
than   the non-Tabu controlled version, no matter the strategy used. Furthermore, we can state that the
 combination Controller + Tabu offers results which are comparable (and even better, see instance 31)
to the CC based GASAT + Tabu.  The comparison amongst GA+control+Tabu strategies and Tabu only shows that 
 the single-steady Tabu Search does not  offer 
 satisfactory results over a broad set of 
 instances (unif*, simon, 3bit and F500): over these instances, a simple 
 comparison with table \ref{mstd} shows that even a non-tabu ReactiveMoving performs better. This allow us to state that 
 the good performances of Tabu + Control are not due just to the Tabu mechanism, as it could be argued:  the adaptive operation 
 selection provide the Tabu search with an efficient way to escape from local optima, with the advantage to be general w.r.t. the instance at hand.

In order to check the robustness of our findings we have defined 20 different sets of operators, each containing an exploitation-oriented operator, an exploration-oriented operator, and 18 randomly chosen ones out of the 300s operators derived from the table \ref{crossoverOps}.
For each operator set, we have rerun experiments using the policies defined above.
For the 20 sets, we have remarked that the Tabu+Control policies perform better than Tabu-only and policies that do not use Tabu improvement.
Adding the Tabu algorithm helps in improving the results of the 
control policies. We can anyhow remark that  non-Tabu policies also provide robust results with regards to the different sets of operators.





Additionally we have run a pairwise Wilcoxon test on the best solutions found by the different policies for each of the 30 rounds found over all instances, in order to verify the  Tabu skill to allow the controlled GA escaping from local optima. 
All possible pairwise combinations amongst 
\begin{itemize}
\item
$TABU+INCREASING,$
\item
$TABU+REACTIVEMOVING,$ 
\item
$TABU+CC,$ 
\end{itemize}
have a p-value greater than $0.05$, leading us to accept the H-hypothesis that 
the distribution from which they are drawn are equivalent. 
Conversely, the tiny p-value found by a pairwise comparisons about each
of the aforementioned strategies and TABU ONLY lead us to confirm that TABU is to be used as a feature to add to a controlled GA 
instead of using a stand-alone strategy.

\section{Conclusions}
\label{conclusions}
 
In this paper we have investigated the  control ability of adaptive control techniques for EAs. The  control consists in achieving  a dynamic adaptation of the algorithm with respect to a given search policy that is defined according to high level criteria, i.e., the quality and the diversity of the population. We have considered various control strategies, in order to handle more dynamic scenarios.

This work has addressed some important aspects related to the automatic control of EAs, namely:
\begin{enumerate}
\item
The ability to identify and select suitable operators for achieving a given search strategy;
\item The ability to maintain a given  search policy by automatically adjusting the EAs' parameters, by means of selecting the operator to apply at each step of the search process;
\item
The ability to solve problems and to perform better than non-controlled EAs.
\end{enumerate}

Results show that dynamic strategies are better than fixed  search policies, in terms of solution quality and 
operators management. Furthermore, the dynamic version allows the EA to better allocate computational time and is more robust w.r.t. the setting of the controller. 
 
The contribution of this paper is thus focused on  providing deep insights for users willing 
to use EAs for solving  specific problems. In this context, adaptive control can be used for two complementary purposes:
\begin{itemize}
\item
Controlling a basic EA in which classic or less known operators have been included  without having any  knowledge about parameters setting. In particular, in presence of many parameters (as in our study, where we consider 20 operators), it is virtually impossible to forecast the impact of the application of these operators during the search, while it would be more intuitive to think in terms of search policy, managing a higher level criterion.
\item
Improving the design of EAs for expert users, for which adaptive control can be used to study the behavior of customised operators according to various search scenarios. 
We have shown that a good controller may achieve good results using ``average'' operators compared to the best performing stand-alone ones, whose design normally requires the execution of costly and time-consuming experiences.
\end{itemize}
 
Further work will be devoted to autonomously modify the operator set during the execution time, and to devise new criteria to define the desired behavior.

\bibliographystyle{elsarticle-harv}
\bibliography{asc}

\begin{thebibliography}{52}
\expandafter\ifx\csname natexlab\endcsname\relax\def\natexlab#1{#1}\fi
\expandafter\ifx\csname url\endcsname\relax
  \def\url#1{\texttt{#1}}\fi
\expandafter\ifx\csname urlprefix\endcsname\relax\def\urlprefix{URL }\fi

\bibitem[{Anton and Olson(2009)}]{sgi}
Anton, C., Olson, L., 2009. Generating satisfiable sat instances using random
  subgraph isomorphism. In: Gao, Y., Japkowicz, N. (Eds.), Advances in
  Artificial Intelligence. Vol. 5549 of Lecture Notes in Computer Science.
  Springer, pp. 16--26.

\bibitem[{Auer(2002)}]{Auer2002b}
Auer, P., 2002. Using confidence bounds for exploitation-exploration
  trade-offs. Journal of Machine Learning Research 3, 397--422.

\bibitem[{Biere et~al.(2009)Biere, Heule, van Maaren, and Walsh}]{Biere2009}
Biere, A., Heule, M., van Maaren, H., Walsh, T. (Eds.), 2009. Handbook of
  Satisfiability. Vol. 185 of Frontiers in Artificial Intelligence and
  Applications. IOS Press.

\bibitem[{Birattari et~al.(2002)Birattari, St\"{u}tzle, Paquete, and
  Varrentrapp}]{BSP02}
Birattari, M., St\"{u}tzle, T., Paquete, L., Varrentrapp, K., 2002. A racing
  algorithm for configuring metaheuristics. In: Proceedings of the Genetic and
  Evolutionary Computation Conference, GECCO. Morgan Kaufmann Publishers Inc.,
  San Francisco, CA, USA, pp. 11--18.

\bibitem[{Burke et~al.(2004)Burke, Gustafson, and Kendall}]{Burke2004}
Burke, E., Gustafson, S., Kendall, G., 2004. Diversity in genetic programming:
  An analysis of measures and correlation with fitness. IEEE Transactions on
  Evolutionary Computation 8~(1), 47--62.

\bibitem[{Burke et~al.(2010)Burke, Hyde, Kendall, Ochoa, Ozcan, and
  Woodward}]{Burke2009a}
Burke, E.~K., Hyde, M., Kendall, G., Ochoa, G., Ozcan, E., Woodward, J., 2010.
  A Classification of Hyper-heuristic Approaches. Vol. 146. Springer US, pp.
  449--468.

\bibitem[{Chatalic and Simon(2000)}]{simon}
Chatalic, P., Simon, L., 2000. Multi-resolution on compressed sets of clauses.
  In: Twelth International Conference on Tools with Artificial Intelligence ,
  ICTAI. pp. 2--10.

\bibitem[{Cheeseman et~al.(1991)Cheeseman, Kanefsky, and Taylor}]{hard}
Cheeseman, P., Kanefsky, B., Taylor, W.~M., 1991. Where the really hard
  problems are. In: Proceedings of {IJCAI}--91. pp. 331--337.

\bibitem[{Cook and Mitchell(1997)}]{Cook1997}
Cook, S., Mitchell, D., 1997. Satisfiability Problem: Theory and Applications.
  DIMACS Series in Discrete Mathematics and Theoretical Computer Science.
  American Mathematical Society, Ch. Finding Hard Instances of the
  Satisfiability Problem: A Survey.

\bibitem[{{Da Costa} et~al.(2008){Da Costa}, Fialho, Schoenauer, and
  Sebag}]{bandit-GECCO08}
{Da Costa}, L., Fialho, A., Schoenauer, M., Sebag, M., 2008. Adaptive operator
  selection with dynamic multi-armed bandits. In: {M. Keijzer et al.} (Ed.),
  Proceedings of the Genetic and Evolutionary Computation Conference, GECCO.
  ACM Press, pp. 913--920.

\bibitem[{di~Tollo et~al.(2011)di~Tollo, Lardeux, Maturana, and
  Saubion}]{Tollo2011}
di~Tollo, G., Lardeux, F., Maturana, J., Saubion, F., 2011. From adaptive to
  more dynamic control in evolutionary algorithms. In: Proceedings of
  Evolutionary Computation in Combinatorial Optimization - 11th European
  Conference, EvoCOP. Vol. 6622 of Lecture Notes in Computer Science. Springer,
  pp. 130--141.

\bibitem[{Eiben et~al.(2007)Eiben, Michalewicz, Schoenauer, and
  Smith}]{eiben06parameterControl}
Eiben, A., Michalewicz, Z., Schoenauer, M., Smith, J., 2007. Parameter Setting
  in Evolutionary Algorithms. Springer, Ch. Parameter Control in Evolutionary
  Algorithms, pp. 19--46.

\bibitem[{Eiben and Smith(2003)}]{eiben03intro}
Eiben, A., Smith, J., 2003. Introduction to Evolutionary Computing. Natural
  Computing Series. Springer.

\bibitem[{Eiben et~al.(1999)Eiben, Hinterding, and
  Michalewicz}]{eiben99parameterControl}
Eiben, A.~E., Hinterding, R., Michalewicz, Z., 1999. Parameter control in
  evolutionary algorithms. IEEE Transaction on Evolutionary Computation 3~(2),
  124--141.

\bibitem[{Eiben and Smit(2012)}]{Eiben2012}
Eiben, A.~E., Smit, S.~K., 2012. Evolutionary algorithm parameters and methods
  to tune them. In: Hamadi, Y., Monfroy, E., Saubion, F. (Eds.), Autonomous
  Search. Springer, pp. 15--36.

\bibitem[{Fialho et~al.(2008)Fialho, {Da Costa}, Schoenauer, and
  Sebag}]{bandit-PPSN08}
Fialho, A., {Da Costa}, L., Schoenauer, M., Sebag, M., 2008. Extreme value
  based adaptive operator selection. In: {G. Rudolph et al.} (Ed.), Parallel
  Problem Solving from Nature - PPSN X, 10th International Conference. Vol.
  5199 of Lecture Notes in Computer Science. Springer, pp. 175--184.

\bibitem[{Fialho et~al.(2010{\natexlab{a}})Fialho, Da~Costa, Schoenauer, and
  Sebag}]{Fialho2010}
Fialho, A., Da~Costa, L., Schoenauer, M., Sebag, M., 2010{\natexlab{a}}.
  Analyzing bandit-based adaptive operator selection mechanisms. Annals of
  Mathematics and Artificial Intelligence 60, 25--64,
  10.1007/s10472-010-9213-y.
\newline\urlprefix\url{http://dx.doi.org/10.1007/s10472-010-9213-y}

\bibitem[{Fialho et~al.(2010{\natexlab{b}})Fialho, Schoenauer, and
  Sebag}]{Fialho2010a}
Fialho, {\'A}., Schoenauer, M., Sebag, M., 2010{\natexlab{b}}. Toward
  comparison-based adaptive operator selection. In: Proceedings of the Genetic
  and Evolutionary Computation Conference, GECCO. ACM, pp. 767--774.

\bibitem[{Fleurent and Ferland(1996)}]{fleurent96}
Fleurent, C., Ferland, J.~A., 1996. Object-oriented implementation of heuristic
  search methods for graph coloring, maximum clique, and satisfiability. In:
  Cliques, Coloring, and Satisfiability: Second DIMACS Implementation
  Challenge. Vol.~26 of DIMACS Series in Discrete Mathematics and Theoretical
  Computer Science. pp. 619--652.

\bibitem[{Glover and Laguna(1999)}]{Glover1999}
Glover, F., Laguna, M., 1999. TABU search. Kluwer.

\bibitem[{Goldberg(1989)}]{Goldberg1989}
Goldberg, D.~E., 1989. Genetic Algorithms in Search, Optimization, and Machine
  Learning. Addison-Wesley.

\bibitem[{Goldberg(1990)}]{GoldbergMatching90}
Goldberg, D.~E., 1990. {Probability Matching, the Magnitude of Reinforcement,
  and Classifier System Bidding}. Machine Learning 5~(4), 407--426.

\bibitem[{Gong et~al.(2010)Gong, Fialho, and Cai}]{Gong2010}
Gong, W., Fialho, {\'A}., Cai, Z., 2010. Adaptive strategy selection in
  differential evolution. In: Genetic and Evolutionary Computation
  Conference,GECCO. ACM, pp. 409--416.

\bibitem[{Hogg(1996)}]{tad}
Hogg, T., 1996. Refining the phase transition in combinatorial search.
  Artificial Inteligence 81, 127--154.

\bibitem[{Holland(1975)}]{Holland1975}
Holland, J.~H., 1975. Adaptation in Natural and Artificial Systems: {A}n
  Introductory Analysis with Applications to Biology, Control and Artificial
  Intelligence. University of Michigan Press.

\bibitem[{Hoos(2012)}]{Hoos2012}
Hoos, H.~H., 2012. Automated algorithm configuration and parameter tuning. In:
  Hamadi, Y., Monfroy, E., Saubion, F. (Eds.), Autonomous Search. Springer
  Berlin Heidelberg, pp. 37--71.

\bibitem[{Hutter et~al.(2007)Hutter, Hoos, and St\"{u}tzle}]{HutHooStu07}
Hutter, F., Hoos, H.~H., St\"{u}tzle, T., 2007. Automatic algorithm
  configuration based on local search. In: Proceedings of the Twenty-Second
  Conference on Artifical Intelligence, AAAI. pp. 1152--1157.

\bibitem[{Lardeux et~al.(2006)Lardeux, Saubion, and Hao}]{Lardeux2006}
Lardeux, F., Saubion, F., Hao, J.-K., 2006. {GASAT}: A genetic local search
  algorithm for the satisfiability problem. Evolutionary Computation 14~(2),
  223--253.

\bibitem[{Linhares and Yanasse(2010)}]{linhares_search_2010}
Linhares, A., Yanasse, H., 2010. Search intensity versus search diversity: a
  false trade off? Applied Intelligence 32~(3), 279--291.

\bibitem[{Lobo et~al.(2007)Lobo, Lima, and Michalewicz}]{Lobo2007}
Lobo, F., Lima, C., Michalewicz, Z. (Eds.), 2007. Parameter Setting in
  Evolutionary Algorithms. Vol.~54 of Studies in Computational Intelligence.
  Springer.

\bibitem[{Lobo and Goldberg(1997)}]{Lobo1997}
Lobo, F.~G., Goldberg, D.~E., 1997. Decision making in a hybrid genetic
  algorithm. In: IEEE International Conference on Evolutionary Computation,
  CEC. IEEE Press, pp. 121--125.

\bibitem[{Maturana et~al.(2012)Maturana, Fialho, Saubion, Schoenauer, Lardeux,
  and Sebag}]{Maturana2012}
Maturana, J., Fialho, A., Saubion, F., Schoenauer, M., Lardeux, F., Sebag, M.,
  2012. Adaptive operator selection and management in evolutionary algorithms.
  In: Hamadi, Y., Monfroy, E., Saubion, F. (Eds.), Autonomous Search. Springer
  Berlin Heidelberg, pp. 161--189.

\bibitem[{Maturana et~al.(2009)Maturana, Fialho, Saubion, Schoenauer, and
  Sebag}]{cec2009}
Maturana, J., Fialho, A., Saubion, F., Schoenauer, M., Sebag, M., 2009. Compass
  and dynamic multi-armed bandits for adaptive operator selection. In:
  Proceedings of IEEE Congress on Evolutionary Computation CEC.

\bibitem[{Maturana et~al.(2010)Maturana, Lardeux, and Saubion}]{Maturana2010}
Maturana, J., Lardeux, F., Saubion, F., 2010. Autonomous operator management
  for evolutionary algorithms. Journal of Heuristics.

\bibitem[{Maturana and Saubion(2008)}]{compass-PPSN08}
Maturana, J., Saubion, F., 2008. A compass to guide genetic algorithms. In: {G.
  Rudolph et al.} (Ed.), Parallel Problem Solving from Nature - PPSN X, 10th
  International Conference. Vol. 5199 of Lecture Notes in Computer Science.
  Springer, pp. 256--265.

\bibitem[{McKay(2000)}]{div1}
McKay, R., 2000. Fitness sharing in genetic programming. In: Proceedings of the
  Genetic and Evolutionary Computation Conference. pp. 435--442.

\bibitem[{Moscato(1989)}]{moscato}
Moscato, P., 1989. On evolution, search, optimization, genetic algorithms and
  martial arts : Towards memetic algorithms. Tech. Rep. C3P 826, Caltech
  Concurrent Computation Program.

\bibitem[{Nannen et~al.(2008)Nannen, Smit, and Eiben}]{Nannen2008}
Nannen, V., Smit, S.~K., Eiben, A.~E., 2008. Costs and benefits of tuning
  parameters of evolutionary algorithms. In: Parallel Problem Solving from
  Nature - PPSN X, 10th International Conference Dortmund, Germany, September
  13-17, 2008, Proceedings. Vol. 5199 of Lecture Notes in Computer Science.
  Springer, pp. 528--538.

\bibitem[{Qin et~al.(2009)Qin, Huang, and Suganthan}]{Qin2009}
Qin, A.~K., Huang, V.~L., Suganthan, P.~N., 2009. Differential evolution
  algorithm with strategy adaptation for global numerical optimization. IEEE
  Trans. Evolutionary Computation 13~(2), 398--417.

\bibitem[{Rivest et~al.(1992)Rivest, Cormen, and Leiserson}]{Rivest1992}
Rivest, R., Cormen, T., Leiserson, C., 1992. Introduction to Algorithms.
  McGraw--Hill, MIT Press, Cambridge, MA.

\bibitem[{Robbins(1952)}]{Robbins1952}
Robbins, H., 1952. Some aspects of the sequential desing of experiments.
  Bulletin American Mathematical Society~(55), 527--535.

\bibitem[{Rodman(1978)}]{Rodman1978}
Rodman, L., 1978. On the many-armed bandit problem. The Annals of Probability
  6~(3), 491--498.
\newline\urlprefix\url{http://www.jstor.org/stable/2243151}

\bibitem[{Smit and Eiben(2009)}]{smit09comparing}
Smit, S., Eiben, G., 2009. Comparing parameter tuning methods for evolutionary
  algorithms. In: Proceedings of the IEEE Congress on Evolutionary Computation,
  CEC.

\bibitem[{Sutton and Barto(1998)}]{Sutton1998}
Sutton, R.~S., Barto, A.~G., 1998. Reinforcement learning: An introduction.
  IEEE Transactions on Neural Networks 9~(5), 1054--1054.

\bibitem[{Sywerda(1989)}]{Sywerda1989}
Sywerda, G., 1989. Uniform crossover in genetic algorithms. In: Proceedings of
  the third international conference on Genetic algorithms. Morgan Kaufmann
  Publishers Inc., pp. 2--9.

\bibitem[{Tang and Wang(2013)}]{Tang2013}
Tang, L., Wang, X., 2013. A hybrid multiobjective evolutionary algorithm for
  multiobjective optimization problems. IEEE Trans. Evolutionary Computation
  17~(1), 20--45.

\bibitem[{Tang et~al.(2014)Tang, Zhao, and Liu}]{Tang2014}
Tang, L., Zhao, Y., Liu, J., 2014. An improved differential evolution algorithm
  for practical dynamic scheduling in steelmaking-continuous casting
  production. IEEE Trans. Evolutionary Computation 18~(2), 209--225.

\bibitem[{Thierens(2005)}]{Thierens2005}
Thierens, D., 2005. An adaptive pursuit strategy for allocating operator
  probabilities. In: Genetic and Evolutionary Computation Conference, GECCO.
  ACM, pp. 1539--1546.

\bibitem[{Thierens(2007)}]{thierens07operatorAllocation}
Thierens, D., 2007. {Adaptive Strategies for Operator Allocation}. In: Lobo,
  F., Lima, C., Michalewicz, Z. (Eds.), Parameter Setting in Evolutionary
  Algorithms. Springer Verlag, pp. 77--90.

\bibitem[{Veerapen et~al.(2012)Veerapen, Maturana, and Saubion}]{Nada2012lion}
Veerapen, N., Maturana, J., Saubion, F., 2012. A comparison of operator utility
  measures for on-line operator selection in local search. In: Learning and
  Intelligent OptimizatioN Conference (LION). pp. 497--502.

\bibitem[{Whitacre et~al.(2006)Whitacre, Pham, and Sarker}]{Whitacre2006}
Whitacre, J.~M., Pham, T., Sarker, R.~A., 2006. Use of statistical outlier
  detection method in adaptive evolutionary algorithms. In: Proceedings of the
  8th annual conference on Genetic and evolutionary computation, GECCO. ACM,
  pp. 1345--1352.

\bibitem[{Zhang and Sanderson(2009)}]{Zhang2009}
Zhang, J., Sanderson, A.~C., 2009. Jade: Adaptive differential evolution with
  optional external archive. IEEE Trans. Evolutionary Computation 13~(5),
  945--958.

\end{thebibliography}

\end{document}